\def\authorBlock{
    Byungjun Kim\textsuperscript{1}\thanks{Equal contribution} \qquad
    Patrick Kwon\textsuperscript{2}\footnotemark[1] \qquad
    Kwangho Lee\textsuperscript{2} \qquad
    Myunggi Lee\textsuperscript{2}
    \and
    Sookwan Han\textsuperscript{1} \qquad
    Daesik Kim\textsuperscript{2} \qquad
    Hanbyul Joo\textsuperscript{1}
    \and
    \textsuperscript{1}Seoul National University \qquad
    \textsuperscript{2}Naver Webtoon AI \\
    {\tt\small \href{https://snuvclab.github.io/chupa}{https://snuvclab.github.io/chupa/}}
}

\newif\ifreview 
\newif\ifarxiv \newcommand{\arxiv}{\arxivtrue}
\newif\ifcamera 
\newif\ifrebuttal 

\arxiv 

\pdfoutput=1
\documentclass[10pt,twocolumn,letterpaper]{article}
\ifreview \usepackage[review]{cvpr} \fi
\ifarxiv \usepackage[pagenumbers]{cvpr} \fi
\ifrebuttal \usepackage[rebuttal]{cvpr} \fi
\ifcamera \usepackage{cvpr} \fi

\usepackage{graphicx}
\usepackage{amsmath}
\usepackage{amssymb}
\usepackage{booktabs}

\usepackage{times}
\usepackage{microtype}
\usepackage{epsfig}
\usepackage[table,xcdraw]{xcolor}
\usepackage{caption}
\usepackage{float}
\usepackage{placeins}
\usepackage{color, colortbl}
\usepackage{stfloats}
\usepackage{enumitem}
\usepackage{tabularx}
\usepackage{xstring}
\usepackage{multirow}
\usepackage{xspace}
\usepackage{cuted} 
\usepackage{url}
\usepackage{subcaption}
\usepackage{xcolor}
\usepackage{mathtools}
\usepackage{enumitem}
\usepackage{float}
\usepackage[accsupp]{axessibility} 
\usepackage[hang,flushmargin]{footmisc}
\usepackage[numbers,sort,compress]{natbib}
\usepackage[normalem]{ulem}  
\usepackage{graphicx}
\usepackage{amssymb}

\ifcamera \usepackage[accsupp]{axessibility} \fi

% ...

\newcommand{\figref}[1]{Fig.~\ref{#1}}
\newcommand{\tabref}[1]{Tab.~\ref{#1}}
\newcommand{\eqnref}[1]{Eq.~(\ref{#1})}
\newcommand{\secref}[1]{Sec.~\ref{#1}}
\newcommand{\sArt}[1]{state-of-the-art∼}

\definecolor{SigmaColor}{rgb}{0.98,0.45,0.0}

\ifarxiv  \fi

\newcommand{\R}[1]{{%
    \textbf{%
        \ifstrequal{#1}{1}{\textcolor{red}{R#1}}{%
        \ifstrequal{#1}{2}{\textcolor{blue}{R#1}}{%
        \ifstrequal{#1}{3}{\textcolor{magenta}{R#1}}{%
        \ifstrequal{#1}{4}{\textcolor{teal}{R#1}}{%
                           \textcolor{cyan}{R#1}%
        }}}}%
    }%
}}

\usepackage{xr-hyper}

\makeatletter
\newcommand*{\addFileDependency}[1]{
  \typeout{(#1)}
  \@addtofilelist{#1}
  \IfFileExists{#1}{}{\typeout{No file #1.}}
}

\makeatother

\usepackage[pagebackref,breaklinks,colorlinks]{hyperref}
\usepackage[capitalize]{cleveref}
\crefname{section}{Sec.}{Secs.}
\crefname{table}{Table}{Tables}
\crefname{figure}{Fig.}{Figs.}

\begin{document}

\title{\textit{Chupa}: Carving 3D Clothed Humans from Skinned Shape Priors \\using 2D Diffusion Probabilistic Models}

\author{\authorBlock}
\maketitle
\begin{strip}
\vspace*{-1.5cm}
\centering
 \includegraphics[width=\textwidth]{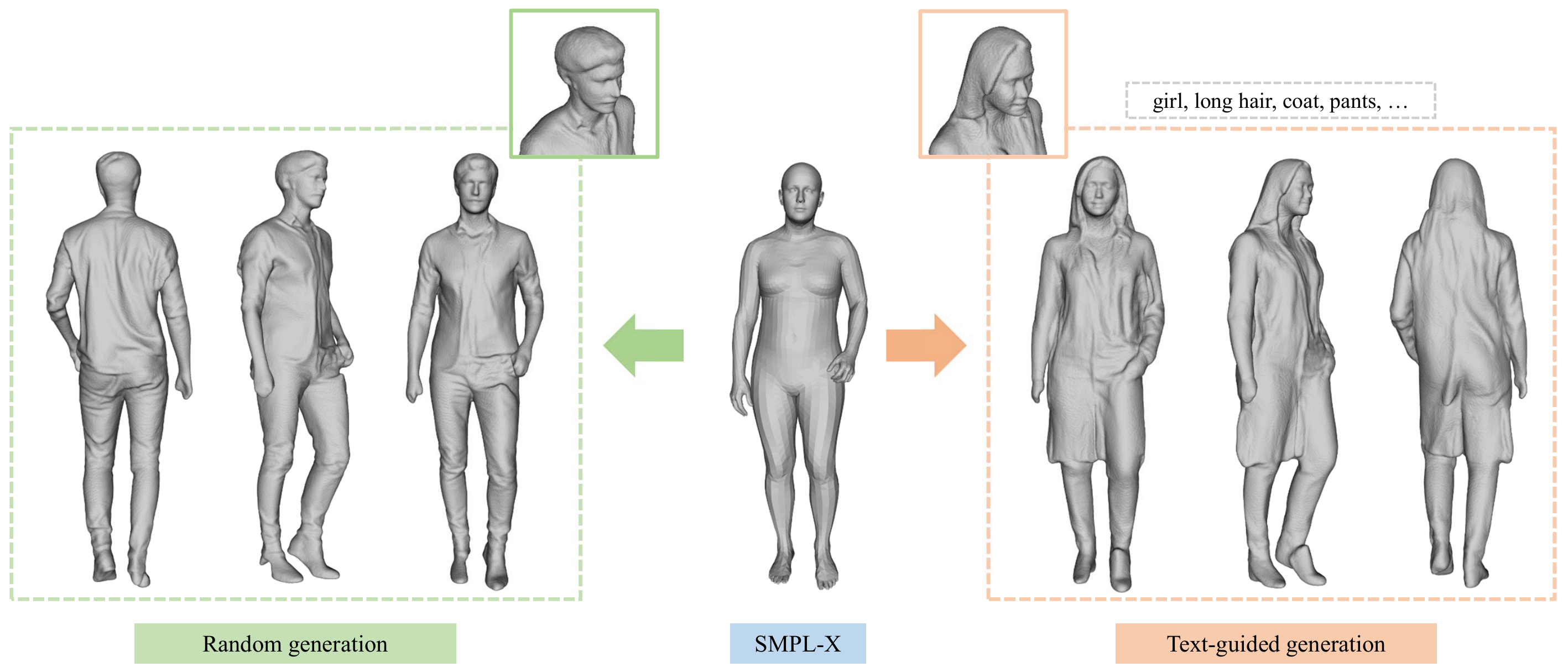}
\captionof{figure}{\textbf{Generative Human Digital Avatars.} We propose \emph{Chupa}, a 3D human generation pipeline that combines the generative power of diffusion models~\cite{Rombach-CVPR-2022ldm} and neural rendering techniques~\cite{Laine-TOG-2020nvdiffrast} to create diverse, and realistic 3D humans. Our pipeline can easily generalize to unseen human poses and display realistic qualities.}
\label{fig:teaser}
\end{strip}

\begin{abstract}
We propose a 3D generation pipeline that uses diffusion models to generate realistic human digital avatars. Due to the wide variety of human identities, poses, and stochastic details, the generation of 3D human meshes has been a challenging problem. To address this, we decompose the problem into 2D normal map generation and normal map-based 3D reconstruction. Specifically, we first simultaneously generate realistic normal maps for the front and backside of a clothed human, dubbed dual normal maps, using a pose-conditional diffusion model. For 3D reconstruction, we ``carve'' the prior SMPL-X mesh to a detailed 3D mesh according to the normal maps through mesh optimization. To further enhance the high-frequency details, we present a diffusion resampling scheme on both body and facial regions, thus encouraging the generation of realistic digital avatars. We also seamlessly incorporate a recent text-to-image diffusion model to support text-based human identity control. Our method, namely, Chupa, is capable of generating realistic 3D clothed humans with better perceptual quality and identity variety.
\end{abstract}
\section{Introduction}
\label{sec:intro}

The creation of clothed 3D human characters, which we refer to as “digital avatars”, has become an essential part of many fields including gaming, animation, virtual/mixed reality, and the 3D industry in general. These digital avatars allow users to use their virtual representation for a range of purposes, thus enhancing user immersion within such services. However, creating high-quality digital avatars often requires specialized 3D artists using a sophisticated creation pipeline~\cite{guo2019relightables, Lombardi2018facerender}, making it a laborious process.

The recent advances in deep generative models \citep{nips2014gan, kingma-ICLR2014-vae, Ho-NIPS-2020ddpm} have enabled the creation of high-quality images that accurately reflects the textual input semantics~\citep{Rombach-CVPR-2022ldm, nichol2022glide}. However, the usage of such generative models in creating 3D has mainly focused on object generation~\citep{Zheng-CGF-2022stylegansdf, Zeng-nips-2022lion, shue20233d, xu2023dream3d, poole-ICLR-2023dreamfusion} and shown rather limited performance in generating full-body, realistic 3D human avatars due to the difficulty of collecting a large-scale ground truth dataset. Many previous 3D generative models~\citep{chan2022efficient, or2022stylesdf, chan2021pi, gu2022stylenerf, zhang2022avatargen, hong2023evad, Bergman-NeurIPS-2022gnarf} focus on training generative models on large-scale image datasets along with implicit 3D shape representations and differentiable volume rendering~\citep{wang2021nerf, niemeyer2020dvr}. However, those approaches are rather limited in generating full-body humans with realistic details and rely on computationally expensive volume rendering. Other approach~\citep{Chen-CVPR-2022gdna} directly uses high-quality 3D datasets~\cite{renderpeople, Yu-CVPR-2021thuman} to train generative models based on auto-decoding frameworks~\cite{park2019deepsdf}, but the resulting stochastic details tend to be unrealistic, due to the usage of an adversarial loss~\citep{nips2014gan}. 

In this paper, we decompose the problem of 3D generation into 2D normal map generation and 3D reconstruction, bridging the power of generative models in the image domain toward 3D generation. Following the intuition of ``sandwich-like'' approaches for single image-based 3D human reconstruction~\cite{gabeur2019moulding, smith2019facsimile, Xiu-CVPR-2023econ}, we generate normal maps for frontal and backside regions of human mesh to get rich details mitigating the computational cost of 3D representations. We adopt a diffusion model~\citep{Ho-NIPS-2020ddpm, Rombach-CVPR-2022ldm} to simultaneously create consistent normal maps for both frontal and backside regions, which we call \emph{dual} normal maps, conditioned on a posed SMPL-X~\citep{Loper-TOG-2015smpl, Pavlakos-CVPR-2019smplx}. Since diffusion models are well known for their mode coverage~\citep{Xiao-ICLR-2022DiffGAN}, we find it suitable to generate diverse 3D digital avatars. The dual normal maps are then used as input for our 3D reconstruction pipeline, in which we \emph{carve} the initial posed SMPL-X mesh to a clothed, realistic human mesh with normal map-based mesh optimization inspired by NDS~\citep{Worchel-CVPR-2022nds}. During optimization, the initial mesh is gradually deformed to match the generated normal maps through a differentiable rasterization pipeline~\citep{Laine-TOG-2020nvdiffrast} and geometric regularization including a loss function for plausible side-view. Our dual normal map-based 3D generation pipeline alleviates the difficulty of generating consistent multi-views, which is the fundamental reason that diffusion-based 3D generative models~\cite{poole-ICLR-2023dreamfusion, wang2023score, xu2023dream3d} suffer from slow convergence or fail to generate multi-view consistent results. We show that the diffusion model can generate consistent dual normal maps and they are sufficient to generate plausible 3D humans along with SMPL-X prior. Then, we can further improve the generated mesh by using a resampling scheme motivated by SDEdit~\cite{Meng-ICLR-2021sdedit}, in which we use separate diffusion models for the body and facial regions to refine the perceptual quality of the rendered normals in different viewpoints while preserving the view and identity consistency. The refined normal maps are subsequently used as inputs for the mesh optimization, thus creating a realistic 3D digital avatar with high-frequency details.

As shown in \figref{fig:teaser}, our pipeline, which we dub it \emph{Chupa}, can be extended to text-based generation for further controllability on the human identity (\eg, gender, clothing, hair, etc.), by leveraging the power of a pre-trained text-to-image diffusion model, e.g., Stable Diffusion~\citep{Rombach-CVPR-2022ldm}. Specifically, we modify and fine-tune the text-to-image model~\citep{brooks2022instructpix2pix, yang2023paintbyexample} to enable conditioning on posed SMPL-X, such that the model creates detailed normal maps according to both the pose information and textual descriptions. Afterward, we pass the generated frontal normal map as guidance to the dual normal map generator to complete dual normal maps, seamlessly connecting text-based generation to our original pipeline.

Trained from posed 3D scans only, Chupa is capable of generating various digital avatars from pose and textual information, with realistic, high-fidelity features such as wrinkles and large varieties in human identity and clothing. We evaluate our method through established benchmarks along with a perceptual study and show that our method outperforms the previous baseline. In summary, our contributions are:
\vspace{-2mm}
\begin{itemize}[itemsep=0pt]
    \item A 3D generation pipeline that directly leverages the 2D image generation capability of diffusion models towards 3D reconstruction.
    \item A diffusion-based normal map generation and refinement strategy for view-consistent normal maps, targeted for 3D generation.
    \item A method to effectively allow text-based 3D full-body digital avatar creation, providing an intuitive scenario for digital avatar creation.
\end{itemize}
\section{Related Work}
\label{sec:related}

\paragraph{3D Generative Models.}
Leveraging the success of generative models in producing realistic 2D images~\cite{nips2014gan, karras2019style, Dhariwal-NIPS-2021guideddiffusion, Karras2019stylegan2, Karras2021stylegan3, fu2022styleganhuman, Fruhstuck-CVPR-2022insetgan}, several efforts have been made to build 3D generative models from 2D datasets while ensuring view consistency~\cite{chan2022efficient, or2022stylesdf, chan2021pi, gu2022stylenerf}. To achieve this, 3D neural implicit representation~\cite{mescheder2019occupancy, park2019deepsdf, wang2021nerf} is employed to represent 3D targets, along with volume rendering to project the 3D scenes into 2D images~\cite{chan2022efficient, or2022stylesdf, chan2021pi, gu2022stylenerf}. While early methods in this direction were mainly focused on rigid objects~\cite{schwarz2020graf, chan2021pi, Niemeyer2020GIRAFFE} or human faces~\cite{chan2022efficient, or2022stylesdf, gu2022stylenerf}, recent work has extended to human bodies by using LBS-based canonicalization~\cite{chen2021snarf} with SMPL to handle articulated pose changes~\cite{zhang2022avatargen, hong2023evad, Bergman-NeurIPS-2022gnarf}. However, these approaches suffer from low-quality 3D outputs and high computational costs due to the volume rendering. 

Other methods~\cite{ma2020learning, corona2021smplicit} utilized SMPL models with latent codes to represent clothing information. However, these methods tend to be limited in geometric detail. gDNA~\citep{Chen-CVPR-2022gdna} was the first generative model-based approach along with a neural implicit representation~\cite{Peng-ECCV-2020convonet} to create diverse 3D humans with varying identities, poses, and clothing. gDNA further leverages the adversarial loss~\citep{nips2014gan} to generate detailed surface normals. However, the adversarial loss made the model susceptible to mode collapse, which leads to unnatural stochastic details. In contrast, our approach is based on diffusion probabilistic models, which alleviates the mode collapsing issue while producing state-of-the-art quality.
    
\begin{figure*}[h] \centering 
    \includegraphics[width=0.9\linewidth,]{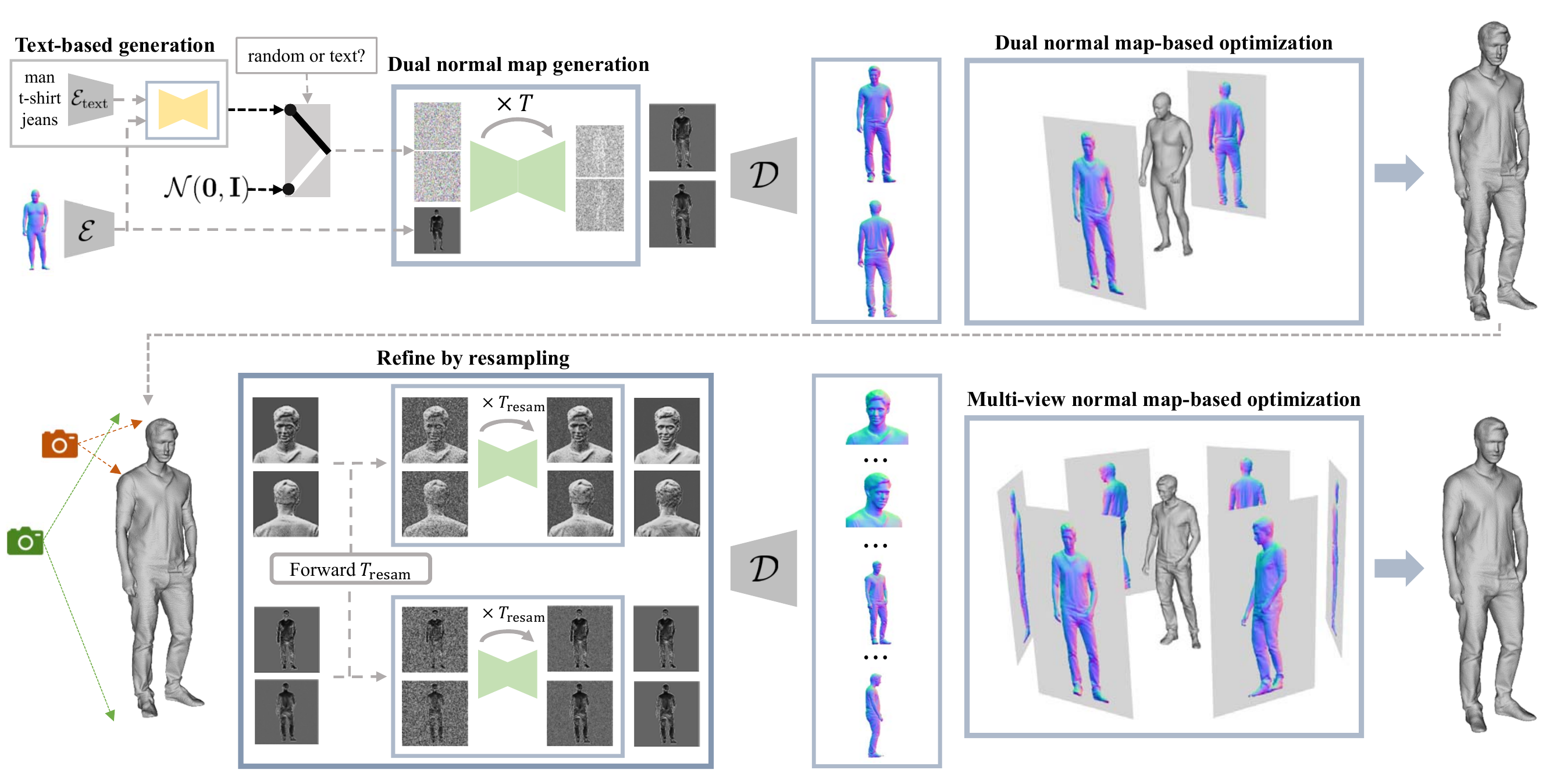}
    \caption{\textbf{Overview.} Chupa takes a posed SMPL-X mesh $\mathcal{M}$ and its front normal map $\mathbf{c}_N$ as input. At the first stage, Chupa generates frontal and backside clothed normal maps, $\mathbf{x}^F, \mathbf{x}^B$, conditioned on $\mathbf{c}_N$. These normals are then used as a reference to ``carve'' $\mathcal{M}$ through our normal map-based mesh optimization process. To further increase the quality, we separately refine the multi-view normal maps rendered from the full body and facial regions through a resampling procedure and perform the second optimization to create $\mathcal{M}_\text{final}$. Our pipeline can also support identity control through a text description by leveraging the power of a text-to-image generation model.}
    \label{fig:pipeline}
\end{figure*}

\paragraph{3D Human Reconstruction.}
The reconstruction of 3D humans has been a long-standing problem in the field of 3D computer vision. Traditional multi-view approaches tended to rely on calibrated multi-camera systems~\cite{Kanade-1997, Matsuyama-2002, Gall-09, Brox-10, Stoll-11, deAguiar-2008, Vlasic-2008, Furukawa-2008, joo2017panoptic}. Several 3D parametric human body models~\cite{anguelov2005scape, Loper-TOG-2015smpl, joo2018, xu2020ghum} have been presented to represent the shape and pose variation of humans through parametric control, and they are widely used in human pose estimation~\cite{kanazawa2018end, kolotouros2019spin, rong2021frankmocap}. Building upon such parametric models, single image-based 3D clothed human reconstruction methods with implicit 3D representation~\cite{Saito-CVPR-2019pifu, Saito-CVPR-2020pifuhd} show outstanding results with high-frequency details. Such models, however, tend to show disembodied or broken limbs for unseen poses due to the lack of topological prior. To address the problem, recent works~\cite{Zheng-TPAMI-2021pamir, Xiu-CVPR-2022icon} combine implicit representation~\cite{mescheder2019occupancy} and parametric models~\cite{Loper-TOG-2015smpl, Pavlakos-CVPR-2019smplx}. Inspired by sandwich-like approaches~\cite{gabeur2019moulding, smith2019facsimile}, ECON~\cite{Xiu-CVPR-2023econ} exploits front and back normal maps to build partial surfaces through normal integration~\cite{cao2022bilateral} and stitches them with a mesh from IF-Net~\cite{chibane20ifnet} and SMPL mesh through poisson surface reconstruction~\cite{kazhdan2006poisson, kazhdan2013screened}. Our approach achieves realistic 3D human generation via normal map-based mesh optimization with SMPL-X mesh as a prior. Rather than using the parametric model as an implicit guidance~\cite{Zheng-TPAMI-2021pamir, Xiu-CVPR-2022icon} or stitching it with separate surfaces~\cite{Xiu-CVPR-2023econ}, we directly deform the SMPL-X mesh to be consistent with the input normal maps, using a differentiable rasterizer~\cite{Laine-TOG-2020nvdiffrast}. 

\paragraph{Diffusion Models.}
Diffusion Probabilistic Models~\cite{Sohl-ICML-2015dpm} are a group of generative models that have achieved state-of-the-art results in perceptual image quality and mode coverage \citep{Dhariwal-NIPS-2021guideddiffusion, ho-2022cascaded, saharia-SIGGRAPH-2022palette, saharia-TPAMI-2022image, Lugmayr-CVPR-2022repaint, Song-NIPS-2019generative}. Recent diffusion models for text-to-image generation~\cite{nichol2022glide, ramesh-arxiv-2022dalle2, Rombach-CVPR-2022ldm, saharia-NIPs2022-photorealistic} have demonstrated the ability to produce high-quality images based on textual input. Among them, \citet{Rombach-CVPR-2022ldm} enhances the efficiency of diffusion models by operating in a latent space that has a lower dimension than the image space while being perceptually equivalent. We list details of the inner workings of the diffusion models in the supplementary material. 

Previous methods~\cite{poole-ICLR-2023dreamfusion, wang2023score, xu2023dream3d, watson-iclr-20233dim} focused on text-to-shape tasks, where the output is a small 3D object lacking photorealistic quality. Among such methods, 3DiM~\citep{watson-iclr-20233dim} presents view-consistent generation through stochastic conditioning but is limited to expressing 3D objects in a $128$ resolution. DiffuStereo~\cite{shao-ECCV-2022diffustereo} was one of the first methods to achieve high-quality 3D human reconstruction through diffusion models, but the usage of diffusion models was limited to refining details, while ours better utilizes the generation capability and mode coverage in generating diverse 3D models. Other work such as Rodin~\cite{wang-arxiv-2022rodin} also uses textual conditions to generate human 3D models, but are limited to the upper body, being unable to represent various human poses.

\section{Method}
\label{sec:method}

Our model is capable of generating 3D full body human models by conditioning on a front normal map rendered from a SMPL-X~\citep{Loper-TOG-2015smpl, Pavlakos-CVPR-2019smplx} mesh $\mathcal{M}$, which provides pose information, and an optional textual description that includes other identity-related information. The resulting 3D clothed human models display realistic details, while maintaining consistency to the input pose and textual description.

Conditioned on the normal map rendered from SMPL-X mesh, we first utilize a diffusion-based generative model to create full body normal maps for both frontal (observed) and backside (occluded) regions (\secref{sec:front_to_back}). We then employ a normal map-based mesh optimization method inspired by NDS~\citep{Worchel-CVPR-2022nds} to deform the posed SMPL-X mesh into a detailed human mesh (\secref{sec:nds}). To enhance the quality of our mesh, we render the normal maps from the resulting human mesh at multiple viewpoints and refine them through a diffusion-based resampling strategy~\cite{Meng-ICLR-2021sdedit}, where we use separate diffusion models for the full body and facial regions (\secref{sec:refine}). The refined normal maps are subsequently used as inputs to our mesh optimization method, creating a high-quality 3D clothed digital avatar. Our pipeline also accepts additional text information to further control the identity of the digital avatar using a text-to-image diffusion model~\citep{Rombach-CVPR-2022ldm} (\secref{sec:text_diffusion}). \figref{fig:pipeline} shows the overall pipeline of our method.

\subsection{Dual Normal Map Generation}
\label{sec:front_to_back}
Following the intuition of ``sandwich-like'' approaches for single image-based 3D human reconstruction~\cite{gabeur2019moulding, smith2019facsimile, Xiu-CVPR-2023econ}, we generate both the frontal and backside normal map $(\mathbf{x}^{F}, \mathbf{x}^{B})$ of clothed humans, dubbed \emph{dual} normal maps, with the front-view SMPL-X normal map $\mathbf{c}_N(\pmb{\beta}, \pmb{\theta})$ as a pose condition, where $\pmb{\beta}, \pmb{\theta}$ are the shape parameters and pose parameters of SMPL-X, respectively. We demonstrate that dual normal maps have sufficient information to generate plausible 3D humans with our normal map-based mesh reconstruction method. By generating dual normal maps, we can mitigate the difficulty and computational cost of directly generating 3D representation (\eg, voxels, point clouds, etc.) or multi-view consistent 2D representation (\eg, RGB images, normal maps, etc.). Since dual normal maps can be represented as images, we can exploit a diffusion model renowned for its image generation capability. We employ a latent diffusion model~\citep{Rombach-CVPR-2022ldm} and adapt it to generate the dual normal maps. Note that we can control the body shape and pose of the generated dual normal maps by changing $\pmb{\beta}, \pmb{\theta}$ with the SMPL-X normal map $\mathbf{c}_N(\pmb{\beta}, \pmb{\theta})$ as a condition. 

Following the latent diffusion model~\cite{Rombach-CVPR-2022ldm}, we first train a vector-quantized autoencoder $(\mathcal{E}, \mathcal{D})$~\citep{Oord-NIPS-2017NeuralDR, Esser-2020-taming} to support normal maps with alpha channels which enable getting foreground mask of generated normal maps easily. Specifically, given a normal map (color-coded as RGB) with alpha channel $\mathbf{x}\in\mathbb{R}^{H{\times}W{\times}4}$, the encoder $\mathcal{E}$ encodes $\mathbf{x}$ into the latent representation $\mathbf{z}\in\mathbb{R}^{h{\times}w{\times}4}$, and the decoder $\mathcal{D}$ reconstructs a normal map back from the latent $\mathbf{z}$. We train our autoencoder based on rendered normal maps from views with different yaw angles so that the autoencoder efficiently encodes these normal maps into a perceptually equivalent latent space, i.e., $\mathbf{z}^F = \mathcal{E}(\mathbf{x}^F)$ and $\mathbf{z}^B = \mathcal{E}(\mathbf{x}^B)$. 
For simultaneous generation, we concatenate the two latent codes $\mathbf{z}^F$ and $\mathbf{z}^B$ into a latent code $\mathbf{z}$ and treat it as an 8-channel image.

\begin{figure}
     \centering
     \begin{subfigure}[b]{0.22\textwidth}
         \centering
         \includegraphics[width=\textwidth]{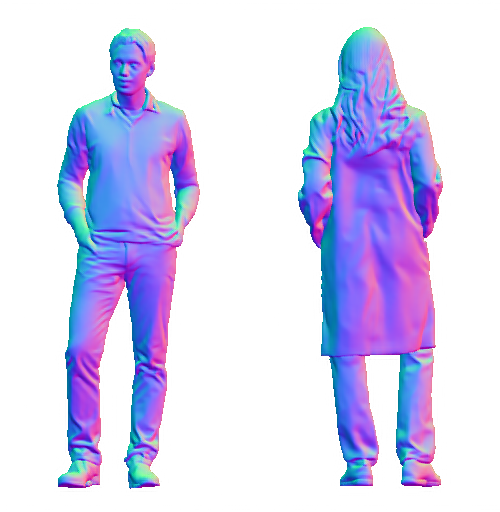}
         \caption{Separate generation}
         \label{fig:body_before_resample}
     \end{subfigure}
     \begin{subfigure}[b]{0.22\textwidth}
         \centering
         \includegraphics[width=\textwidth]{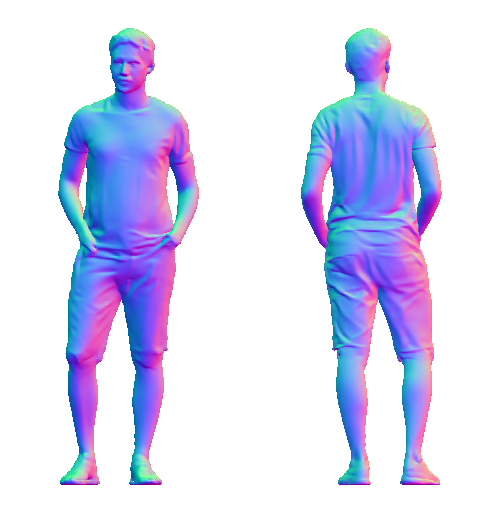}
         \caption{Dual generation}
         \label{fig:body_after_resample}
     \end{subfigure}
     \hfill
    \caption{\textbf{Separate generation vs. Dual generation.} Comparison between (a) separate sampling for frontal/backside normal maps and (b) our dual sampling. When generated separately, attributes of two normal maps likely differ. However, generating the dual normal maps at once ensures the maps share the same semantics.} 
\label{fig:single_vs_dual}
\end{figure}

During training, the latent code $\mathbf{z}$ is perturbed by the forward diffusion process according to a timestep $t$, producing a noisy latent code $\mathbf{z}_t$. The diffusion model $\pmb{\epsilon}_\theta$ then learns to predict the perturbed noise $\pmb{\epsilon}$ of $\mathbf{z}_t$, given the SMPL-X normal map condition $\mathbf{c}_N(\pmb{\beta}, \pmb{\theta})\in\mathbb{R}^{H{\times}W{\times}4}$. 
which is also encoded into $\mathcal{E}(\mathbf{c}_N)\in\mathbb{R}^{h{\times}w{\times}4}$ and concatenated with $\mathbf{z}_t$ channelwise. 
The corresponding objective becomes
\begin{equation}\label{eq:ldm_loss_2}
 L_{\text{dual}} =\mathbb{E}_{\mathbf{x}^F, \mathbf{x}^B, \mathbf{c}_N, \pmb{\epsilon}
 \sim\mathcal{N}(\mathbf{0}, \mathbf{I}), t}[\|\pmb{\epsilon}-\pmb{\epsilon}_\theta(\mathbf{z}^F_t, \mathbf{z}^B_t, t, \mathcal{E}(\mathbf{c}_N))\|^2_2].
\end{equation}
At inference time, we start from the Gaussian noise $\mathbf{z}_T \sim \mathcal{N}(\mathbf{0}, \mathbf{I})$ and iteratively sample from the previous step until $\mathbf{z}_0$, then we decode $\mathbf{z}_0$ to get the final frontal and backside normal maps. We use classifier-free guidance~\citep{Ho-arxiv-2022CFG} to boost the sample quality during conditional generation. To enable classifier-free guidance, we randomly assign blank latent embeddings to the conditional image $\mathbf{c}_N$ with $10$\% probability during training. Then, for each inference step, we use the following modification to predict the denoised latent code:
\begin{equation}\label{eq:cfg_eps}
    \hat{\pmb{\epsilon}}_\theta(\mathbf{z}_t, t, \mathcal{E}(\mathbf{c}_N)) = \lambda \pmb{\epsilon}_\theta(\mathbf{z}_t, t, \mathcal{E}(\mathbf{c}_N)) + (1 - \lambda)\pmb{\epsilon}_\theta(\mathbf{z}_t, t),
\end{equation}
where $\lambda$ specifies the guidance strength that can be controlled during inference, and $\pmb{\epsilon}_\theta(\mathbf{z}_t, t, \mathcal{E}(\mathbf{c}_N))$ and $\pmb{\epsilon}_\theta(\mathbf{z}_t, t)$ each corresponds to the conditional and unconditional predictions. In \figref{fig:single_vs_dual}, our simultaneous dual generation scheme shows that the generated frontal and backside normal maps are more consistent, compared to separate generation.

\subsection{Mesh Reconstruction with Front/Back Normals}
\label{sec:nds}

Given the initial posed SMPL-X mesh $\mathcal{M(\pmb{\beta}, \pmb{\theta}})$ and the generated clothed normal maps $(\mathbf{x}^F, \mathbf{x}^B)$, we deform the initial mesh into a detailed 3D human mesh through iterative optimization. Our mesh reconstruction method is motivated by Neural Deferred Shading (NDS)~\citep{Worchel-CVPR-2022nds}, which reconstructs geometry from multi-view RGB images using a differentiable rasterizer and neural shader. Unlike NDS, we remove the neural shader as the generated normal maps provide supervision for geometry, and directly optimize the 3D geometry by comparing the normal maps with the geometry buffers rendered from a differentiable rasterizer~\citep{Laine-TOG-2020nvdiffrast}. In general, mesh reconstruction via the two normal maps is an ill-posed problem due to the depth ambiguity. Using SMPL-X mesh as an initial mesh, which is a strong geometric prior, and introducing a novel side loss $L_{\mathrm{sides}}$ for regularizing side-views, we can reconstruct plausible 3D geometry of humans while mitigating the difficulty of generating multi-view consistent images at once. Our total objective is defined as
\begin{equation}
\begin{gathered}
L=\lambda_{\mathrm{normal}}L_{\mathrm{normal}}+\lambda_{\mathrm{mask}}L_{\mathrm{mask}}+\lambda_{\mathrm{sides}}L_{\mathrm{sides}}\\+\lambda_{\mathrm{laplacian}}L_{\mathrm{laplacian}}+\lambda^{\mathrm{reg}}_{\mathrm{normal}}L^{\mathrm{reg}}_{\mathrm{normal}}.
\end{gathered}
\end{equation}
\paragraph{Normal map loss.} We minimize the difference between the input normal maps $(\mathbf{x}^F, \mathbf{x}^B)$ and the normal maps rendered from the front/back views of the human mesh $(\mathbf{N}^F, \mathbf{N}^B)$ through a $L_1$ loss, denoted as $L_{\mathrm{normal}}$. We also minimize the discrepancy between the mask of the normal maps through a $L_2$ loss, $L_{\mathrm{mask}}$, to match the silhouette of the mesh. Note that we can acquire the masks of the generated normal maps by a simple thresholding on the alpha channel.

\paragraph{Side loss.}
Since our initial 3D reconstruction is based on frontal/backside normal maps, the left/rightside regions of the human body tend to contain depth ambiguity~\cite{Queau-JMIV-2018normal}. We therefore introduce a novel side loss, which ensures that the body masks rendered from the side views $(\hat{\mathbf{M}}_\text{left}, \hat{\mathbf{M}}_\text{right})$ are not shrinked into the side views of the initial SMPL-X mesh $(\mathbf{M}_{\text{left}}^{\text{smpl}}, \mathbf{M}_{\text{right}}^{\text{smpl}})$. The loss function becomes
\begin{equation}\label{eq:sides}
\begin{gathered}
    L_{\mathrm{sides}}=\sum_{\mathbf{M}^{\mathrm{smpl}}_\mathrm{view}[h, w] = 1
    } \|{\mathbf{M}^{\mathrm{smpl}}_\mathrm{view}[h, w] - \hat{\mathbf{M}}_\mathrm{view}[h, w]}\|^{2}_{2},
\end{gathered}
\end{equation}
where $[h, w]$ denotes indexing with the pixel $(h, w)$ of the mask $\mathbf{M}\in\mathbb{R}^{H{\times}W}$ and $\text{view}\in\{\text{left}, \text{right}\}$. Even though we can mitigate the problem to some extent with the 3D prior from initial SMPL-X, we further prevent the optimized mesh from having unrealistic side-views.

\paragraph{Geometric regularization.}
As noted by NDS~\cite{Worchel-CVPR-2022nds}, optimizing the mesh based on only the aforementioned loss terms can lead to degenerated mesh due to unconstrained vertex movement. To overcome this issue, we use geometric regularization terms following NDS~\cite{Worchel-CVPR-2022nds}. Given a matrix $\mathbf{V}\in{\mathbb{R}^{n\times3}}$ with vertex positions of mesh $\mathcal{M}$ as rows, the Laplacian term is defined as $L_{\mathrm{laplacian}}=\frac{1}{n}\sum_{i=1}^{n}{\|\pmb{\delta}_{i}\|_{2}^{2}}$,
where $\pmb{\delta}_{i}=(LV)_{i}\in{\mathbb{R}^3}$ are the differential coordinates of vertex $i$ with the graph Laplacian $L$. Since the differential coordinates are the sum of positional difference between its neighbors, minimizing this loss leads to a smoother mesh. We also introduce a normal consistency term, defined as $L^{\mathrm{reg}}_{\mathrm{normal}}=\frac{1}{|\bar{\mathcal{F}}|}\sum_{(i, j)\in\bar{\mathcal{F}}}(1-{\pmb{n}}_{i} {\cdot} {\pmb{n}_{j}})^{2}$,
where $\bar{\mathcal{F}}$ is the set of mesh face pairs with a shared edge and $\pmb{n}_{i}\in\mathbb{R}^3$ is the normal of triangle $i$. Minimizing the cosine similarity between face normals of neighbors encourages further smoothness.
\begin{figure}
     \centering
     \begin{subfigure}[b]{0.2\textwidth}
         \centering
         \includegraphics[width=\textwidth]{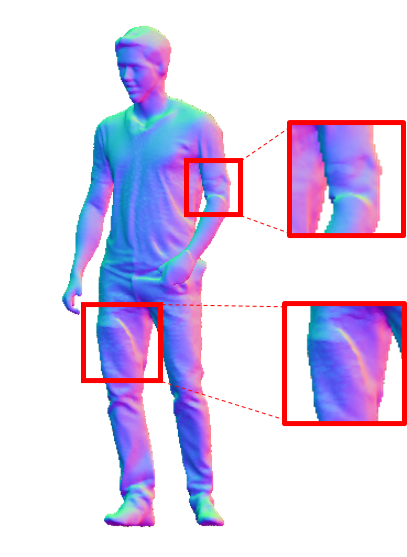}
         \caption{Rendered normal}
         \label{fig:body_before_resample}
     \end{subfigure}
     \begin{subfigure}[b]{0.2\textwidth}
         \centering
         \includegraphics[width=\textwidth]{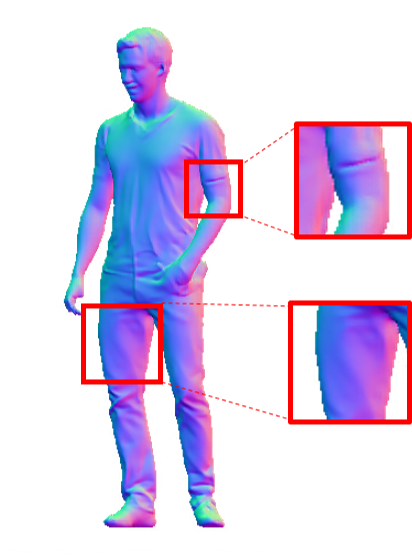}
         \caption{Resampled normal}
         \label{fig:body_after_resample}
     \end{subfigure}
     \hfill
    \caption{\textbf{Body Resampling}. The initial 3D mesh displays undesired visual artifacts, such as unnatural cloth wrinkles and depth misprediction. By resampling, those artifacts are moderated to produce more natural results.} 
\label{fig:body_resampling}
\end{figure}

\subsection{Refine by Resampling}
\label{sec:refine}
\paragraph{Resampling multi-view normal maps.}

After the initial mesh reconstruction, we can further improve the mesh while we already have plausible one. We refine the 3D human mesh by refining the rendered multi-view normal maps of the reconstructed mesh without losing view consistency. The refined maps are then used as inputs to the 3D reconstruction pipeline, creating an improved, realistic 3D human mesh.

Our pipeline is inspired by SDEdit~\citep{Meng-ICLR-2021sdedit}, which proposes an image translation method by progressively denoising a noise-perturbed image. The amount of noise perturbation is decided by timestep $0 < t_0 < 1$, and as $t_0$ gets closer to 0, the operation focuses on editing the finer details. We repeat this process by $K$ times to improve fidelity without harming the original information. To preserve the original structure while adjusting any unrealistic information, we set $t_0 = 0.02$ and $K=2$, which we empirically found to be sufficient. 

In practice, we first render a collection of $n$-view normal maps $\{ \mathbf{I}^1, \mathbf{I}^2, ..., \mathbf{I}^{n} \}$ by evenly rotating the yaw camera angle around the 3D mesh. For refinement, we use the same dual normal map generation model in \secref{sec:front_to_back}, which uses the normal map of posed SMPL-X as spatial guidance. We pair the rendered normal maps so that each pair is rendered from the backside of one another, and use the SMPL-X normal map corresponding to the frontal normal map as the condition to the diffusion model. This perturb-and-denoise process, which we call \emph{resampling}, drives the normal maps rendered from the optimized mesh into the distribution of normal maps rendered from training 3D scans on which our diffusion model is trained, thus the normal maps become more realistic without losing overall semantics. Once the resampling is complete, we pass the refined normal maps as inputs to the 3D reconstruction stage (\secref{sec:nds}) to produce a refined 3D human model. \figref{fig:body_resampling} shows that our resampling-based refinement produces more natural details.

\begin{figure}
\centering
     \begin{subfigure}[b]{0.2\textwidth}
         \centering
         \includegraphics[width=\textwidth]{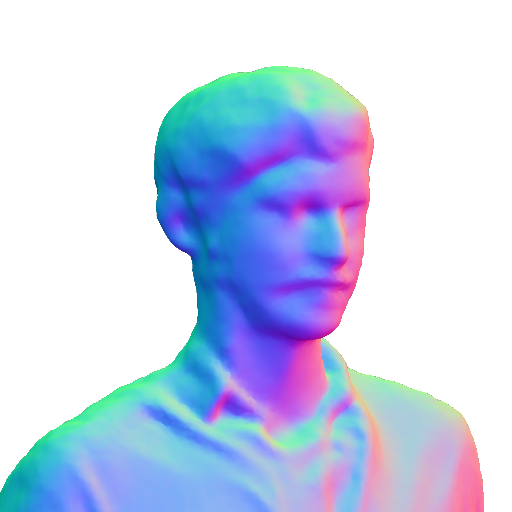}
         \caption{Rendered normal}
         \label{fig:face_before_resample}
     \end{subfigure}
     \begin{subfigure}[b]{0.2\textwidth}
         \centering
         \includegraphics[width=\textwidth]{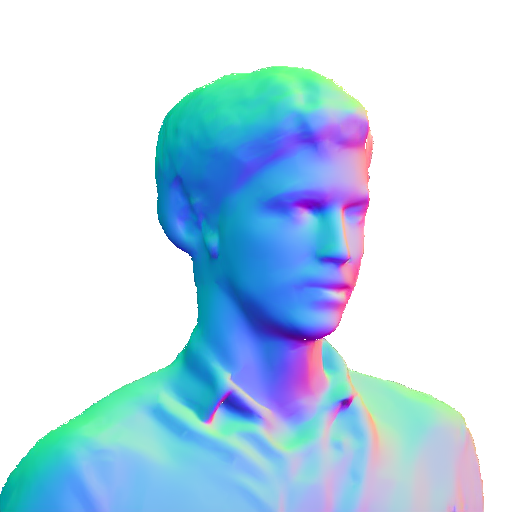}
         \caption{Resampled normal}
         \label{fig:face_after_resaple}
     \end{subfigure}
     \hfill
     \caption{\textbf{Face close-up resampling}. Both images are aligned according to the SMPL-X vertices for the facial region. We can observe that the perceptibility of the faces is improved.} 
\label{fig:face_resampling}
\end{figure}

\paragraph{Facial resampling.}
We enhance the facial details of the optimized mesh by refining the normal maps rendered from the facial regions of the mesh. We train a latent diffusion model which shares the same architecture of the dual normal map generation model in \secref{sec:front_to_back}, but trained on normal maps with face close-up. The close-up is done for the head vertices of SMPL-X based on the pre-defined part segmentation~\cite{Pavlakos-CVPR-2019smplx}. With the face close-up views, we can render facial regions of 3D scans and aligned SMPL-X mesh.

Given the aligned facial normal maps, we can train the diffusion model which generates the frontal and backside facial normal maps with facial normal maps of SMPL-X as a condition. We then apply the same resampling technique used for the full body to refine the multi-view facial normal maps rendered from the optimized mesh. \figref{fig:face_resampling} shows how the facial region is perceptually refined without harming the original structure. Unlike the method of \citet{Fruhstuck-CVPR-2022insetgan}, which performs offline optimization to blend a full body image and face image, we just do the normal map-based optimization (\secref{sec:nds}) with refined normal maps of both body and face, which aggregates the refined normal maps directly in 3D to generate a 3D human mesh with better details. 

\subsection{Text-guided Normal Map Generation}
\label{sec:text_diffusion}

In addition to the main, pose-conditional 3D generation pipeline, we also include an optional pose-and-text conditional pipeline to further control the identity of the resulting human mesh. To generate 3D human mesh based on a textual description, we adopt a powerful text-to-image diffusion model, \eg, Stable Diffusion~\citep{Rombach-CVPR-2022ldm}, and fine-tune its weights to generate normal maps that are consistent to the text description and the posed SMPL-X normal map.

As the method of \citet{Wang-arxiv-2022PITI} displayed the effectiveness of fine-tuning large diffusion models for image translation tasks, we initialize the weights of our model based on a pre-trained Stable Diffusion checkpoint, leveraging its renowned generation capabilities. Following previous works ~\cite{yang2023paintbyexample, brooks2022instructpix2pix}, we add additional input channels to the first layer of the U-Net~\cite{ronneberger-miccai2015-unet} and initialize their weights to zero. We also use the same text conditioning based on a pre-trained CLIP model~\cite{radford-ICML2021-clip}. 

\begin{figure}
     \centering
     \begin{subfigure}[b]{0.125\textwidth}
         \centering
         \includegraphics[width=\textwidth]{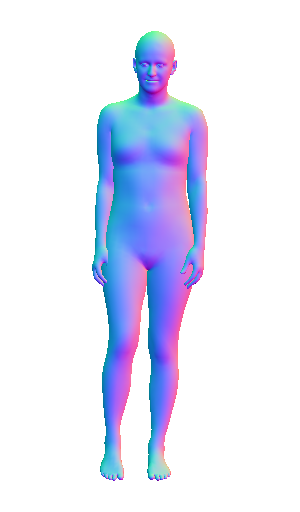}
         \caption{SMPL-X}
         \label{fig:body_before_resample}
     \end{subfigure}
     \hfill
     \begin{subfigure}[b]{0.125\textwidth}
         \centering
         \includegraphics[width=\textwidth]{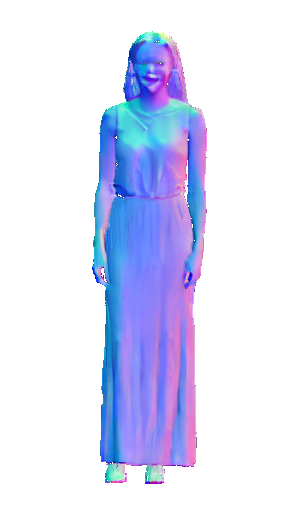}
         \caption{Text-based front}
         \label{fig:body_after_text}
     \end{subfigure}
     \hfill
     \begin{subfigure}[b]{0.172\textwidth}
         \centering
         \includegraphics[width=\textwidth]{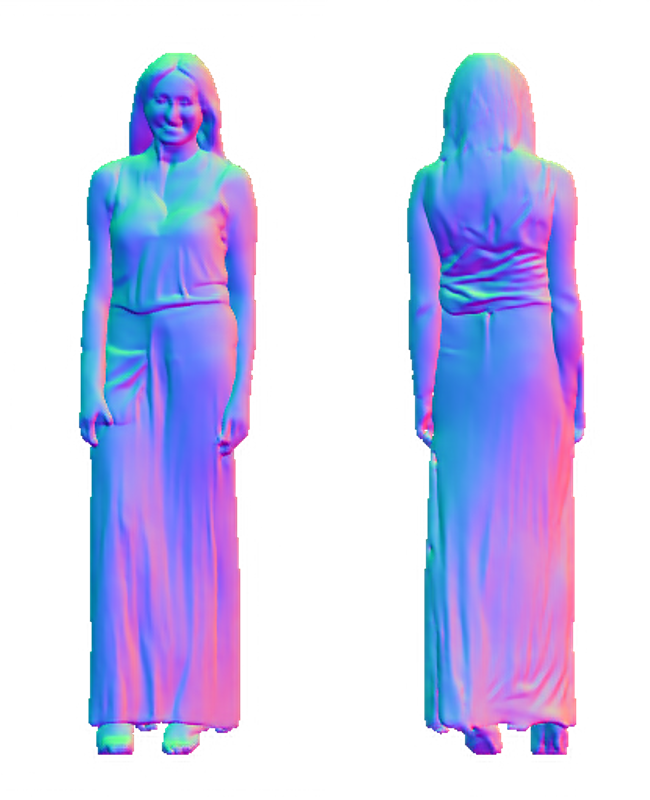}
         \caption{Dual from (b)}
         \label{fig:body_after_resample}
     \end{subfigure}
     \hfill
    \caption{\textbf{Text-based normal map generation.} Note that our model is capable of generating a normal map consistent in gender, clothing, and hair style\protect\footnotemark{}. Moreover, our guided generation method can create a view-consistent back normal map from the initial frontal map, making it possible to use it for our original pipeline.}
\label{fig:text2dual}
\end{figure}

As shown in \figref{fig:text2dual}, our model supports the generation of detailed normal maps based on the textual description and the posed SMPL-X. Our method is the first method to support text-based full-body normal map generation by basing on Stable Diffusion.

\footnotetext{For text, we used "girl, long hair, dress"}

\paragraph{Frontal normal map-guided generation.} To get \emph{dual} normal maps based on the frontal normal map generated from the text-based normal map generation model, we follow the intuitions of Repaint~\cite{Lugmayr-CVPR-2022repaint}. Since we already know and want to preserve the frontal shape, the goal here is to predict the unknown backside normal map, based on the frontal normal map. For each inference step, we sample the intermediate frontal latent code $\mathbf{z}_t^F$ from the original latent $\mathbf{z}^F$ at any timestep $t$, since the diffusion process is defined by a Gaussian Markov chain. In contrast, we sample the unknown, intermediate backside latent code $\mathbf{z}_t^B$ through reverse diffusion, which is concatenated channel-wise to $z_t^F$.  
Since we consider both $\mathbf{z}_t^F$ and $\mathbf{z}_t^B$ as a single, 8-channel latent code, the diffusion model leverages the context of the known frontal normal map while generating the unknown backside normal map, making this a \textit{channel-wise inpainting approach}. \figref{fig:text2dual} shows that our approach helps to generate backside normal maps that match the original frontal map. Through frontal normal map-guided dual normal map generation, we can seamlessly connect the generative powers of a text-to-image model with our main pipeline.
\section{Experiments}
\label{sec:experiments}

\begin{figure*} \centering 
    \includegraphics[width=0.9\linewidth, trim={0 0 0 0},clip]{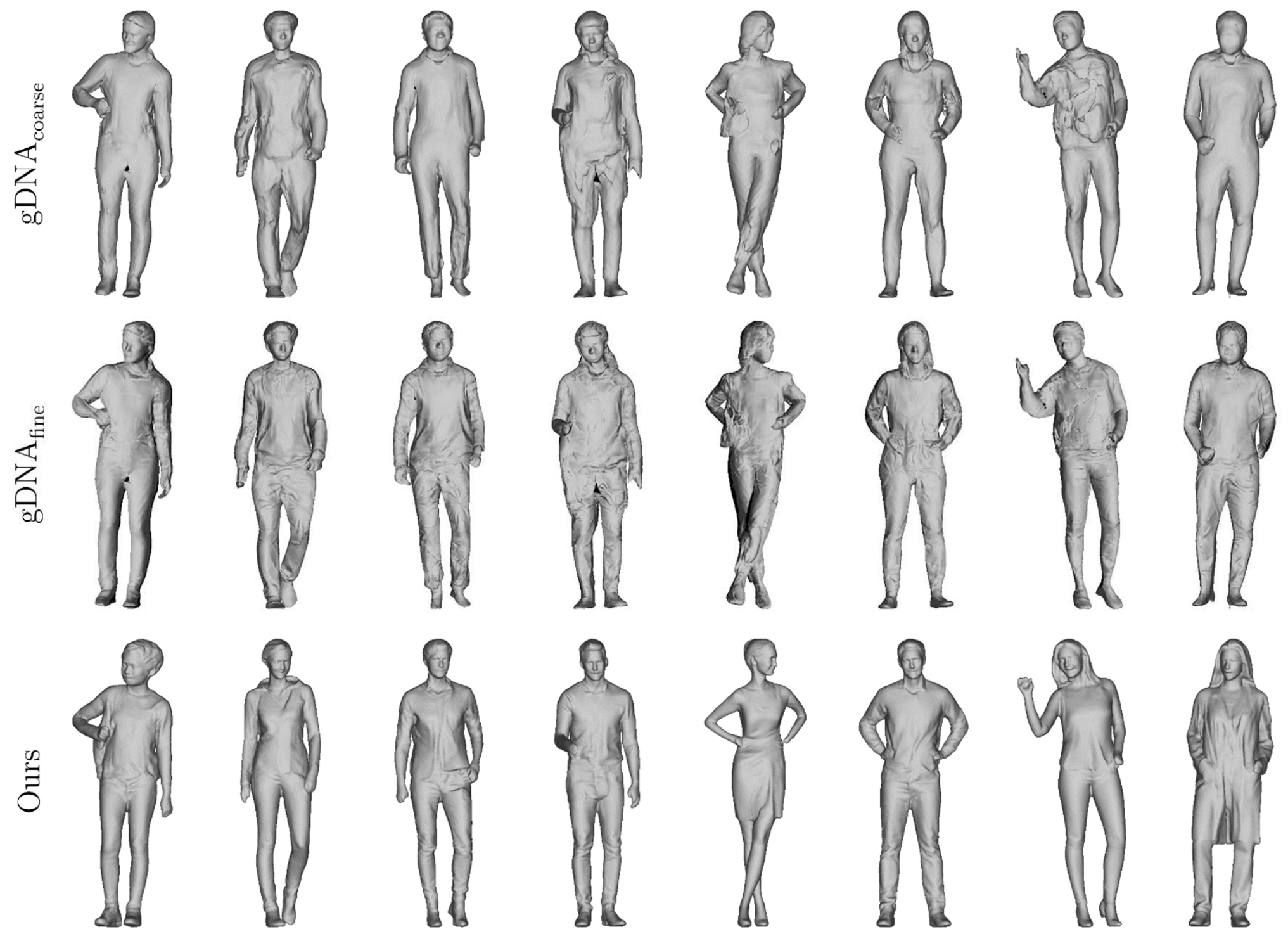}
    \caption{\textbf{Generation Comparison}. We display the visual comparisons between gDNA \citep{Chen-CVPR-2022gdna} and Chupa with the same SMPL input. Note that gDNA tends to amplify the unnatural artifacts from its coarse stage to the fine stage, while our results produce more natural results.} 
\label{fig:compare_gdna}

\end{figure*}

In this section, we validate Chupa's effectiveness in generating realistic 3D humans. We first compare Chupa with the previous state-of-the-art through an image quality metric and a perceptual user study. We also conduct ablation studies to illustrate the effectiveness of each part of our pipeline. \figref{fig:compare_gdna} shows comparison of generated results from our method and the baseline~\cite{Chen-CVPR-2022gdna}. 

\paragraph{Datasets.}
We train and test our model with Renderpeople~\cite{renderpeople} and THuman 2.0~\cite{Yu-CVPR-2021thuman} dataset, which consists of $500$, $526$ scans with various identities and clothing. We split both datasets with a 9:1 ratio for train/test split. For training, we render $36$ multi-view normal maps of the train split scans with rotation of $10^{\circ}$ yaw interval. We follow ICON~\cite{Xiu-CVPR-2022icon} for rendering pipeline, originally from  MonoPort~\cite{li-ECCV-2020monoport}, both for body and face. For rendering normal maps of facial regions, we use the pre-defined part segmentation label of SMPL-X~\cite{Pavlakos-CVPR-2019smplx} to find head vertices of fitted SMPL-X. Then, we render the facial region of 3D scans and fitted SMPL-X mesh with a weak perspective camera for rendering the head vertices of SMPL-X mesh with close-up. To create text pairs from normal maps for Stable Diffusion fine-tuning, we adopt an off-the-shelf image tagger model~\citep{WD14Tagger} based on ViT~\citep{dosovitskiy-iclr-2021vit}. 

\paragraph{Baseline.} 
We compare our method with gDNA~\cite{Chen-CVPR-2022gdna} as a baseline. gDNA is the state-of-the-art method to generate 3D human mesh with given SMPL-X parameter $\beta$, $\Theta$ and randomly sampled shape latent code $\mathbf{z}_{\mathrm{shape}}$ and detail latent code $\mathbf{z}_{\mathrm{detail}}$ from its learned latent space.

\subsection{Implementation Details}
\label{sec:implementation_details}

\paragraph{Autoencoder model training.} Before training the full-body dual generation model, we trained the autoencoder model $(\mathcal{E}, \mathcal{D})$ for $1,000$ epochs on $4 \times$ NVIDIA A100 GPUs following the original implementation \citep{Esser-2020-taming}. We used a VQ-regularized autoencoder with downsampling factor $f = 4$ and channel dimension $c = 4$ such that, given a full-body normal map image with alpha transparency $(\mathbf{c}_N \in \mathbb{R}^{512 \times 512 \times 4})$, the encoder transforms the image to a latent code with 4 channels $(\mathcal{E}(\mathbf{c}_N) \in \mathbb{R}^{128 \times 128 \times 4})$, and the decoder reconstructs the image from the latent code. For training, we used the full-body normal map datasets, following the same preprocessing listed in the main paper. We used the pretrained weights for the autoencoders of facial generation models (\secref{sec:refine}) and text-based generation models (Sec. 3.4) provided by the original paper \citep{Rombach-CVPR-2022ldm}. For the facial generation model, we used a VQ-regularized autoencoder with downsampling factor $f = 4$ and channel dimension $c = 3$. For textual generation models, we used a KL-regularized autoencoder with downsampling factor $f = 8$ and channel dimension $c = 4$. All autoencoders were frozen during diffusion training.

\paragraph{U-net.} We adapt the U-Net~\cite{ronneberger-miccai2015-unet} architecture for our diffusion models to support our dual-generation scheme. Specifically, we follow the approach of \citet{Dhariwal-NIPS-2021guideddiffusion} to further improve the sampling quality and set the input channels from $6$ to $12$, and the output channels from $3$ to $8$. By utilizing the concatenation of two input images (front and back) with the SMPL latent code $\mathcal{E}(\mathbf{c}_N)$ for conditioning, we can treat them as a single input. As a result, we can obtain two spatially aligned images for both views at the same time. For the facial generation models, we set the input channels to $9$ and output channels to $6$, since we used 3-channel for facial normal maps. 

\paragraph{Dual normal map generator training.} We train our full-body dual normal map generation model for $500$ epochs with batch size $16$ on $4 \times$ NVIDIA A100 GPUs. 
We set the total timesteps $T = 100$ with a linear variance schedule. During inference, we use the same $512 \times 512$ resolution and generate results with the same denoising steps used during training. We trained the facial generation model for $300$ epochs with the same training settings.

\paragraph{Text-guided normal map generator training.} We train our text-based normal map generator for $1,000$ epochs on $4 \times$ NVIDIA A100 GPUs. We train at a $512 \times 512$ resolution with a total batch size of 64. We initialize our model from the EMA weights of the Stable Diffusion \citep{Rombach-CVPR-2022ldm} checkpoints and adopt other training settings from the public Stable Diffusion code base. After inference, we used a thresholding operation on the 3rd channel of the image to create a transparency map before the dual generation stage.

\subsection{Quantitative Results}
\label{sec:quantitative}

\begin{table}[t]
\caption{
\textbf{Quantitative Evaluation.} We report two types of FID scores for the test split of Renderpeople and Thuman 2.0.
} \centering
\begin{tabular}{lcc}
\toprule
Method & FID$_{\mathrm{normal}}\downarrow$ & FID$_{\mathrm{shade}}\downarrow$ \\ \hline
\midrule
gDNA$_{\mathrm{coarse}}$~\cite{Chen-CVPR-2022gdna} & $53.74$ & $68.14$\\
gDNA$_{\mathrm{fine}}$~\cite{Chen-CVPR-2022gdna} & $36.43$ & $45.57$ \\
Ours & $\mathbf{21.90}$ & $\mathbf{36.58}$ \\
\bottomrule
\end{tabular}
\vspace{-0.05in}
\label{tab:image_metric_rpth}
\end{table}


We conduct a quantitative evaluation of the quality of generated meshes, based on given SMPL-X parameters.  We generated 3D human meshes with SMPL-X parameters fitted to $103$ test scans, \ie $50$ from Renderpeople and $53$ from THuman $2.0$, for both our method and gDNA~\cite{Chen-CVPR-2022gdna}. Following the previous work~\cite{Chen-CVPR-2022gdna, Zheng-CGF-2022stylegansdf, Shue-arxiv-20223dNDF}, we render normal maps~\cite{Chen-CVPR-2022gdna} and shading-images~\cite{Zheng-CGF-2022stylegansdf, shue20233d} of groundtruth scans and generated meshes into $18$ views with $20^\circ$ yaw interval, and compute FID score with them, which denoted as FID$_{\mathrm{normal}}$ and FID$_{\mathrm{shade}}$ respectively. \tabref{tab:image_metric_rpth} shows that our method achieves lower FID for both images than the baseline.

\subsection{User Preference} 
\label{sec:user_preference}
We carry out a perceptual study over 78 subjects asking about their preference between the meshes from our method and gDNA. We randomly select $40$ from a set of SMPL-X parameters fitted to $103$ test scans. We randomly generate meshes based on them with our method and gDNA, and render shading-images in $3$ views, $0{^\circ}, 120{^\circ}, 240{^\circ}$ for full body images and $0{^\circ}, 40{^\circ}, -40{^\circ}$ for face images. Note that we use the narrower field-of-view for better comparing facial details. \tabref{tab:image_metric_rpth} shows that the users preferred meshes from our method both for full-body and face images. We present more details in the supplementary material.

\begin{table}[t]
\caption{
\textbf{User preference}. We carry out a perceptual study asking 78 subjects to choose a more realistic one between ours and gDNA$_{\mathrm{fine}}$.
}
\centering
\begin{tabular}{lccc}
\toprule
Method &  Body & Face & Total \\ \hline
\midrule
gDNA$_{\mathrm{fine}}$ & $20.89\%$ & $18.7\%$ & $19.78\%$ \\
Ours & $\mathbf{79.11}\%$ & $\mathbf{81.3}\%$ & $\mathbf{80.22}\%$ \\
\bottomrule
\end{tabular}
\vspace{-0.05in}
\label{tab:user_preference}
\end{table}
\begin{table}[t]
\caption{
\textbf{Ablation study.} We do ablation study over our key components. We report FID$_{\mathrm{normal}}$ score.
} \centering
\begin{tabular}{ccccc}
\toprule
dual. & $L_{\mathrm{sides}}$ & refine$_{\mathrm{body}}$ & refine$_{\mathrm{face}}$ & FID$_{\mathrm{normal}}\downarrow$ \\
\midrule
 & & & & $30.55$\\
\checkmark & & & & $26.31$ \\
\checkmark & \checkmark & & & $25.50$ \\
\checkmark & \checkmark & \checkmark & & $22.61$ \\
\checkmark & \checkmark & \checkmark & \checkmark & $\mathbf{21.90}$ \\
\bottomrule
\end{tabular}
\vspace{-0.05in}
\label{tab:image_metric}
\end{table}


\subsection{Ablation Study}
\label{sec:ablation}
We validate the building blocks of our pipeline through an ablation study. The evaluation is based on the same test split. The results are summarized in \tabref{tab:image_metric}. 

\paragraph{Front/Back normal map generation.}
To validate the effectiveness of our dual normal map generation method, we separately generate frontal and backside normal maps with the SMPL normal map in the corresponding view. Due to the randomness of the diffusion model, we cannot guarantee the separately generated frontal and backward normal maps are consistent (\figref{fig:single_vs_dual}), which leads to performance loss.

\begin{figure}
     \centering
     \begin{subfigure}[b]{0.9\columnwidth}
         \centering
         \includegraphics[width=\textwidth]{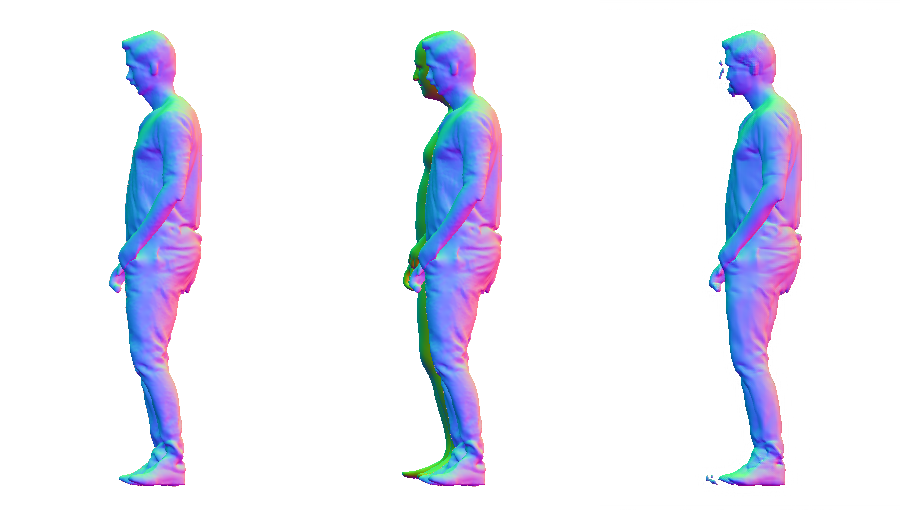}
         \caption{without $L_{\mathrm{sides}}$}
         \label{fig:without_side}
     \end{subfigure}
     \begin{subfigure}[b]{0.9\columnwidth}
         \centering
         \includegraphics[width=\textwidth]{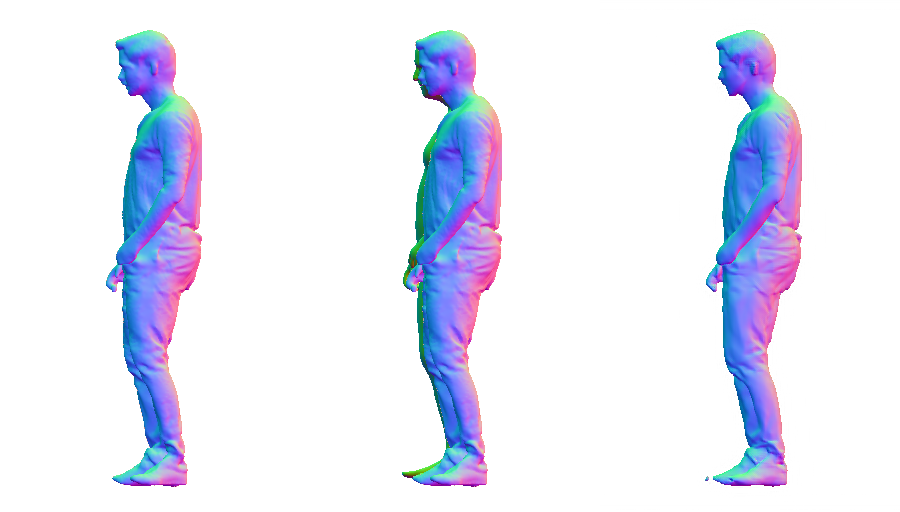}
         \caption{with $L_{\mathrm{sides}}$}
         \label{fig:with_side}
     \end{subfigure}
     \hfill
    \caption{\textbf{Side loss.} We present the side-view normal maps of the optimized mesh (left), the normal maps overlapped on the SMPL-X normal maps (middle), and the normal maps after resampling (right). Without $L_{\mathrm{sides}}$, the alignment between the SMPL-X mesh and the optimized mesh becomes worse, leading to artifacts on the resampling result. (Note that the blue channel of the overlapped SMPL-X normal map is flipped for visualization purposes.)} 
\label{fig:side_loss}
\end{figure}

\paragraph{Side loss.}
With the sidewise loss $L_{\mathrm{sides}}$ from \eqnref{eq:sides}, we enforce our mesh to keep better alignment with the SMPL-X prior during mesh optimization (\secref{sec:nds}). \figref{fig:side_loss} shows the effect of utilizing $L_{\mathrm{sides}}$. The first column shows the side-view normal map rendered from the mesh optimized with dual normal maps. The second column shows the same side-view normal map but overlapped with the side-view of the corresponding SMPL-X. The third column shows the normal maps after resampling (\secref{sec:refine}). \figref{fig:without_side} shows that the optimized mesh without $L_{\mathrm{sides}}$ has worse alignment with SMPL-X mesh, which leads to the artifacts on resampling results. \tabref{tab:image_metric} demonstrates the inclusion of $L_{\mathrm{sides}}$ leads to lower FID scores, indicating its effectiveness.

\paragraph{Refinement.}
To validate the effectiveness of our refinement method (\secref{sec:refine}), we compare 3D generation results only optimized by front/back normal maps and the results refined by body refinement and additional face refinement. \figref{fig:body_resampling} and \figref{fig:face_resampling} show that our refinement methods lead to more realistic generation results. As expected, \tabref{tab:image_metric} shows that our face refinement method further reduces FID.

\section{Discussion}
\label{sec:conclusion}

We propose Chupa, a powerful 3D generation pipeline for a large variety of dressed 3D high-quality digital avatars. By combining diffusion models for normal map generation with a normal map-based mesh reconstruction method, our pipeline enables the creation of realistic 3D avatars with high levels of stochastic details. 
We also allow the creation of 3D humans from both pose and textual information, providing an intuitive method of digital avatar creation. 

We note that while our pipeline can support text conditioning without losing visual quality, several elements that can be generated from the initial text-to-image model (e.g., bracelet, necklace, glasses) tend to be lost during the later stage of the pipeline and cannot be expressed at the final 3D model. For future work, we look forward to creating digital avatars with photorealistic textures and devising novel strategies for creating animations from our digital avatars.
\paragraph{Acknowledgements.} This work was supported by Naver Webtoon. The work of SNU members was also supported by SNU Creative-Pioneering Researchers Program, NRF grant funded by the Korean government (MSIT) (No. 2022R1A2C2092724), and IITP grant funded by the Korean government (MSIT) (No.2022-0-00156 and No.2021-0-01343). H. Joo is the corresponding author.

{\small
\setlength{\bibsep}{0pt}
\bibliographystyle{abbrvnat}
\bibliography{shortstrings, 11_references}
}

\ifarxiv \clearpage \appendix

\label{sec:appendix}
\section{Detailed formulation of Diffusion Models}
\label{sec:supp_diffusion}

We provide a detailed introduction to Gaussian-based diffusion models \citep{Sohl-ICML-2015dpm, Ho-NIPS-2020ddpm}. Given the target data distribution $x_0 \sim q(x_0)$, the goal of diffusion models is to learn a model distribution $p_\theta$ that approximates $q$, while being easy to sample from. To achieve both objectives, diffusion models define a \textit{forward process} that gradually introduces noise to the original data $x_0$ to generate a sequence of noised data $x_1, x_2, ..., x_T$. Additionally, a \textit{reverse process} is defined, which aims to denoise the noised data $x_t$ and produce less noisy data $x_{t-1}$. Once trained, Gaussian-based diffusion models sample data $x_0$ by first sampling $x_T$ from a Gaussian distribution $\mathcal{N}(0, \mathbf{I})$ and iteratively sampling $x_{t-1}$ from the previous step $x_t$. To ensure $x_T \sim \mathcal{N}(0, \mathbf{I})$, it is required for $T$ to be sufficiently large.

The forward process is formulated as a Markov chain according to a variance schedule $\beta_1 < \beta_2 < ... < \beta_T$: 

\begin{equation}
q(x_t | x_{t-1}) := \mathcal{N}(x_t; \sqrt{1-\beta_t}x_{t-1}, \beta_t \mathbf{I})
\end{equation}

\begin{equation}
q(x_{1:T} | x_0) := \prod_{t=1}^T q(x_t | x_{t-1})
\end{equation}
Note that to sample $x_t \sim q(x_t|x_0)$, it is not required to apply forward diffusion $t$ times. Instead, using the notation $\alpha_t := 1 - \beta_t$ and $\bar{\alpha}_t := \prod_{s=1}^t \alpha_s$, we have a closed form expression:

\begin{equation}
q(x_t|x_0) := \mathcal{N}(x_t; \sqrt{\bar{\alpha}_t}x_0, (1-\bar{\alpha}_t\mathbf{I})
\end{equation}
Consequently, we can view $x_t$ as a linear combination of $x_0$ and $\epsilon \sim \mathcal{N}. (0, \mathbf{I})$$(x_t = \sqrt{\bar{\alpha}_t}x_0 + \sqrt{(1 - \bar{\alpha}_t)}\epsilon)$

Given the fixed forward process,  $p$ is designed to approximate the unknown true posterior $q(x_{t-1}|x_t)$. This is achieved through the use of a deep neural network with learnable parameters $\theta$.

\begin{equation}
    p_\theta(x_{t-1}|x_t) := \mathcal{N}(x_{t-1}; \mu_\theta(x_t, t), \Sigma_\theta(x_t, t)) 
\end{equation}

\begin{equation}
    p_\theta(x_{0:T}) := p(x_T)\prod_{t=1}^T p_\theta(x_{t-1} | x_t) 
\end{equation}

\citet{Ho-NIPS-2020ddpm} proposed a specific parameterization for $\mu_\theta(x_t,t)$ such that the neural network outputs the estimated noise $\epsilon_\theta$ instead of predicting $\mu_\theta$.

\begin{equation}
    \mu_\theta(x_t, t) = \frac{1}{\sqrt{\alpha_t}}(x_t - \frac{1-\alpha_t}{\sqrt{1-\bar{\alpha}_t}} \epsilon_\theta(x_t, t))
\end{equation}

For training, the variational lower bound is optimized and simplifies the following \eqnref{eq:lower_bound} that enables the model to learn how to predict the added noise.
\begin{equation} \label{eq:lower_bound}
    L_{\text{simple}} = \mathbb{E}_{x_0,t,\epsilon \sim \mathcal{N}(0,I)} [ || \epsilon - \epsilon_\theta(x_t, t) ||^2 ]
\end{equation}

In practice, \citet{Ho-NIPS-2020ddpm} uses a U-Net backbone \citep{ronneberger-miccai2015-unet} to output the predicted noise $\epsilon_\theta$ which has the same dimensionality as the input noisy sample $x_t$. To solve an image-to-image translation task, \citet{saharia-TPAMI-2022image} concatenates a spatial conditioning input $y$ to $x_t$ channel-wise and modifies the learning objective as \eqnref{eq:cond_diff}.

\begin{equation} \label{eq:cond_diff}
    L_{\text{simple}} = \mathbb{E}_{x_0,y,t,\epsilon \sim \mathcal{N}(0,I)} [ || \epsilon - \epsilon_\theta(x_t, y, t) ||^2 ]
\end{equation}

\begin{figure*}
     \centering
     \begin{subfigure}[b]{0.19\textwidth}
         \centering
         \includegraphics[width=\textwidth]{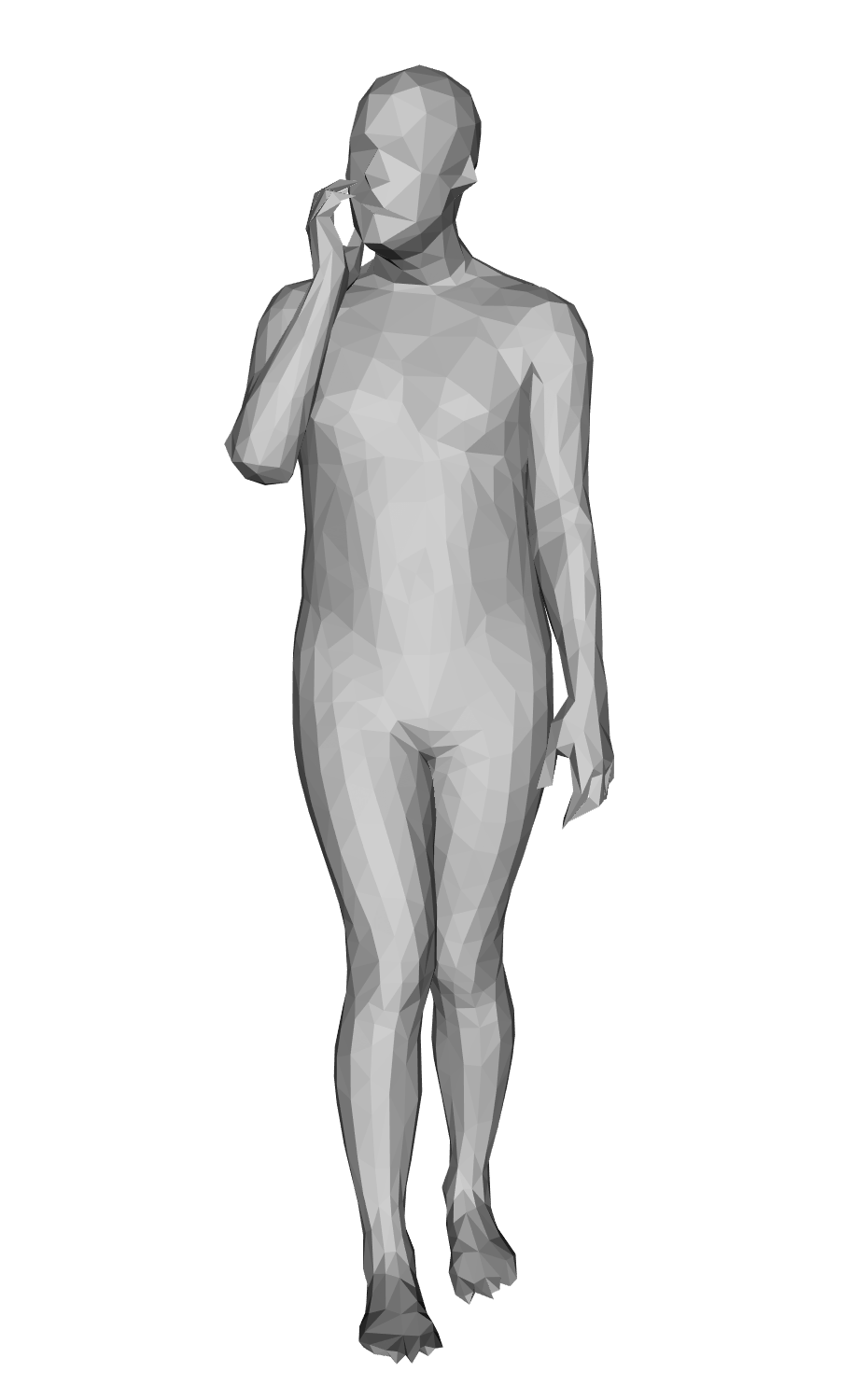}
         \caption{iter. $0$}
         \label{fig:coarse_to_fine_0001}
     \end{subfigure}
     \begin{subfigure}[b]{0.19\textwidth}
         \centering
         \includegraphics[width=\textwidth]{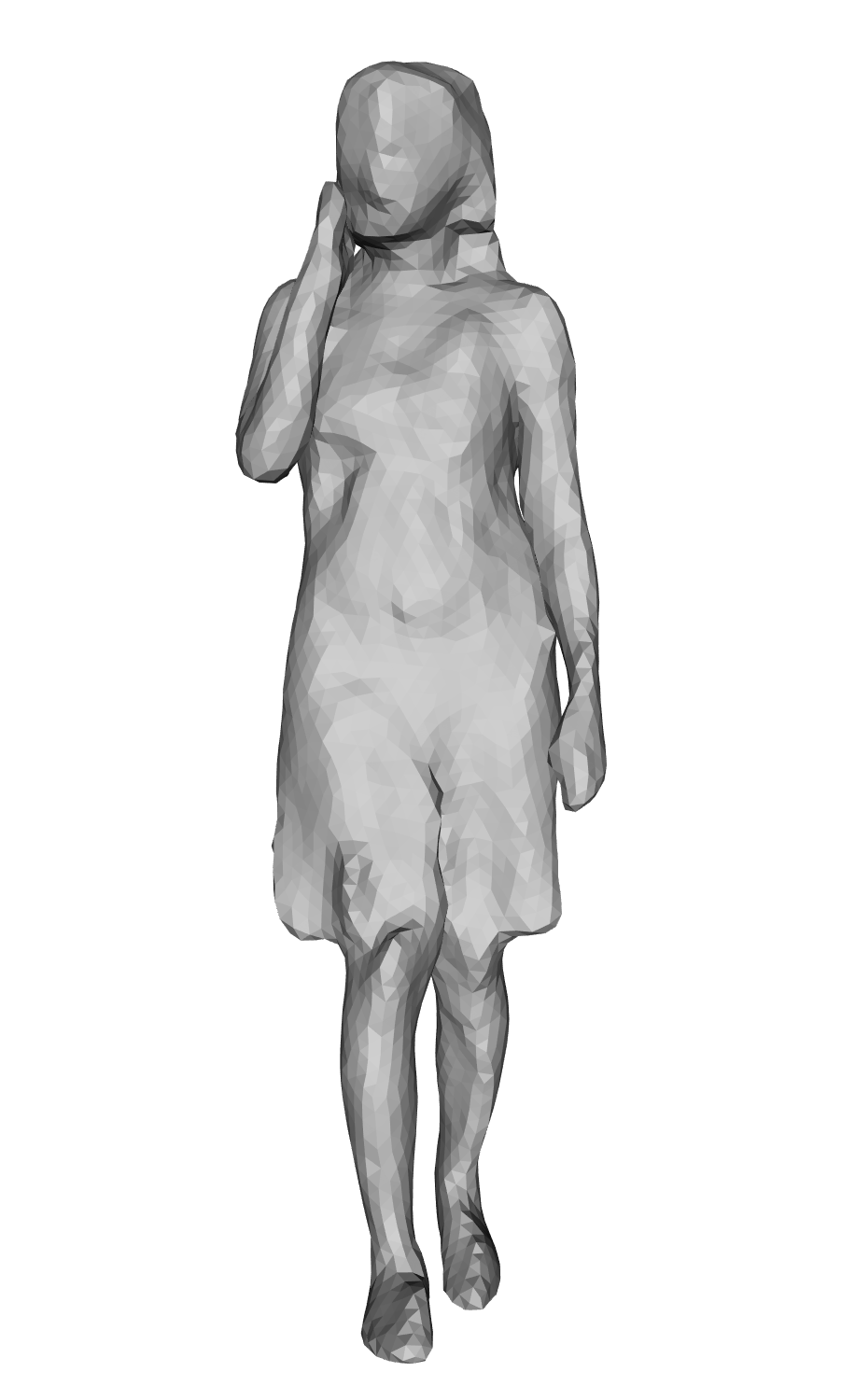}
         \caption{iter. $500$}
         \label{fig:coarse_to_fine_0500}
     \end{subfigure}
     \begin{subfigure}[b]{0.19\textwidth}
         \centering
         \includegraphics[width=\textwidth]{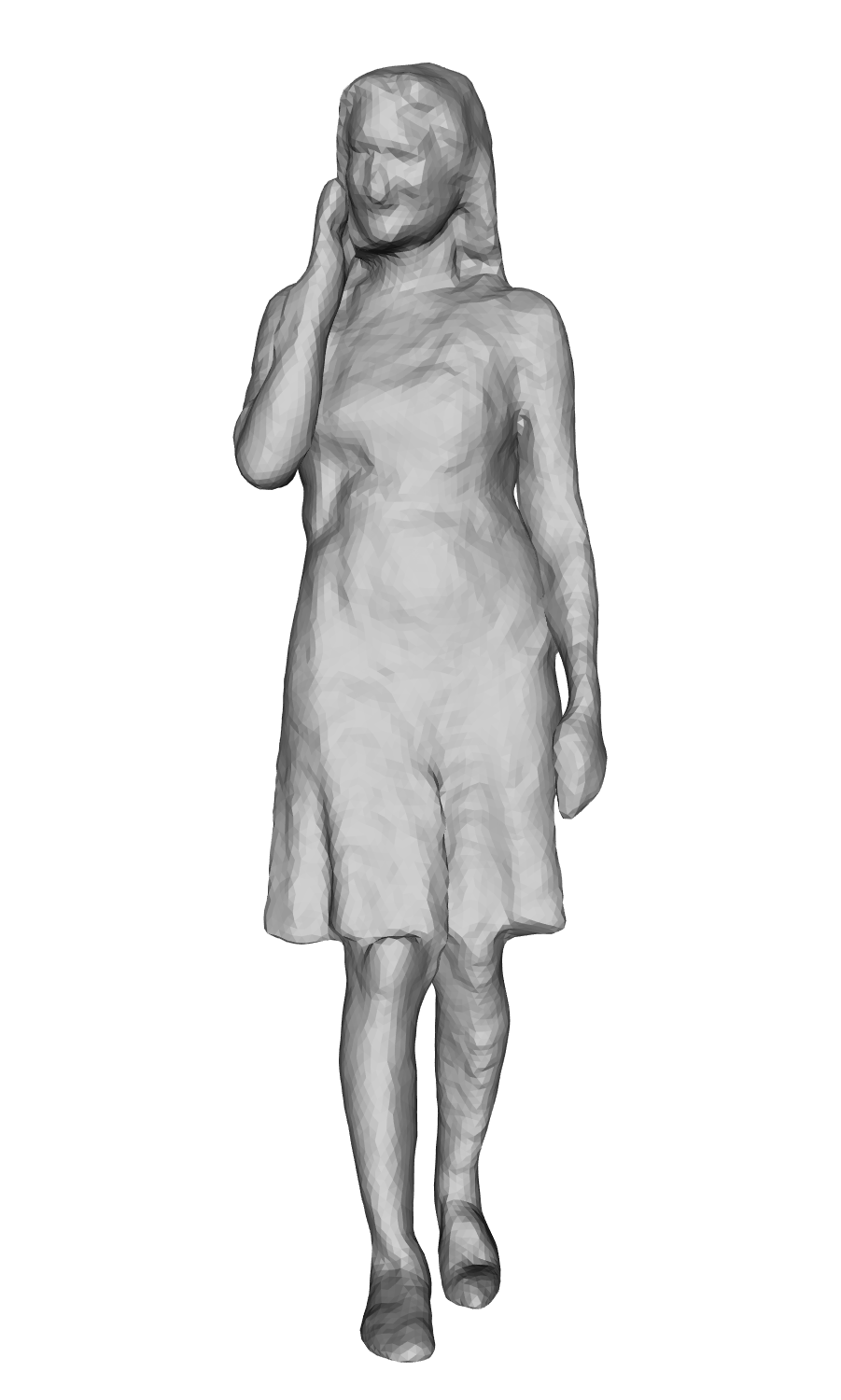}
         \caption{iter. $1000$}
         \label{fig:coarse_to_fine_1000}
     \end{subfigure}
     \begin{subfigure}[b]{0.19\textwidth}
         \centering
         \includegraphics[width=\textwidth]{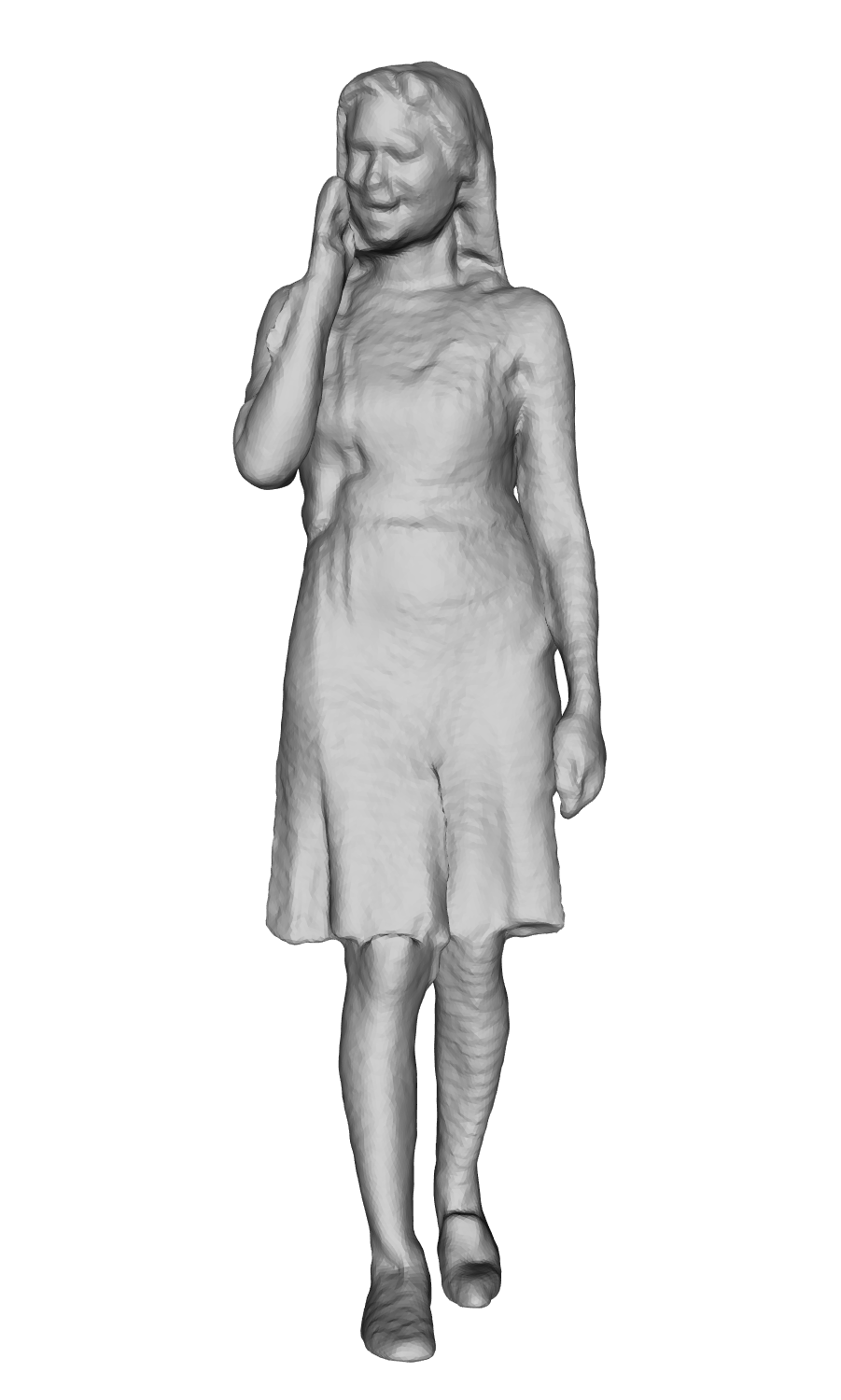}
         \caption{iter. $1500$}
         \label{fig:coarse_to_fine_1500}
     \end{subfigure}
     \begin{subfigure}[b]{0.19\textwidth}
         \centering
         \includegraphics[width=\textwidth]{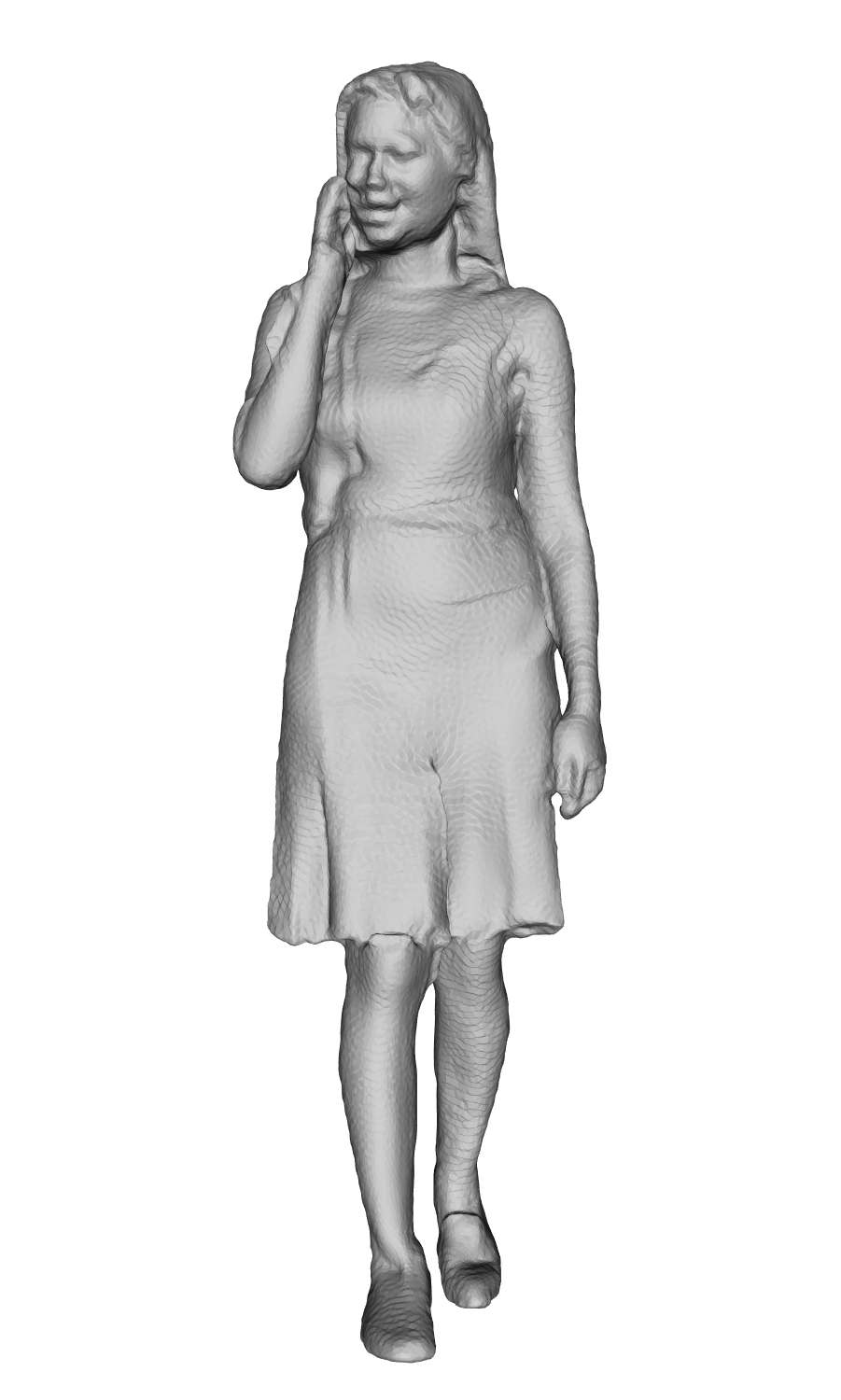}
         \caption{iter. $2000$}
         \label{fig:coarse_to_fine_2000}
     \end{subfigure}
     \hfill
    \caption{\textbf{Coarse-to-fine optimization}. Starting from a decimated SMPL-X mesh, we perform optimization in a coarse-to-fine manner. By increasing the resolution of the mesh for every $500$ iterations, we progressively deform the mesh to match the input normal maps, without losing high-frequency details.} 
\label{fig:coarse-to-fine}
\end{figure*}

\subsection{Diffusion Training}
\label{sec:diffusion_training}

\section{Normal map-based mesh optimization}
\label{sec:supp_optimization}

\paragraph{Camera parameters.}
In our normal map-based mesh optimization method, we require camera parameters to rasterize the mesh into normal maps that are aligned with those generated from our dual-generation diffusion model. To generate the frontal normal map of the initial SMPL-X mesh (explained in Sec. \textcolor{red}{3.1}), we utilize a weak perspective camera which shares the same parameters as our training data setup. For the second mesh refinement stage (explained in Sec. \textcolor{red}{3.3}), we also employ weak perspective cameras that are defined in the same manner for both body and face rendering.

\paragraph{Coarse-to-fine optimization.}
We adopt the coarse-to-fine optimization strategy presented by NDS \citep{Worchel-CVPR-2022nds} for mesh optimization. Specifically, we begin with a coarse mesh and progressively increase the resolution through a remeshing technique, presented by \citet{Botsch-eurographics-2004remeshing}. As demonstrated in \citep{Worchel-CVPR-2022nds}, initializing optimization with a large number of vertices can lead to meshes with undesired geometry, such as degenerate triangles and self-intersections. Therefore, we start the optimization from a decimated version of our initial SMPL-X, which contains 3,000 vertices \citep{Garland-siggraph-1997quadric}. During optimization, for every 500 iterations, we apply remeshing \citep{Botsch-eurographics-2004remeshing} to increase the model resolution. It is worth noting that each iteration corresponds to a single gradient descent step, with respect to the loss based on a randomly sampled normal map. Following NDS \citep{Worchel-CVPR-2022nds}, we perform optimization for a total of 2,000 iterations and decreased the gradient descent step size for the vertices by 25\% after each remeshing. As \figref{fig:coarse-to-fine} shows, we can handle the large deviation from the initial mesh without losing high-frequency details, due to the coarse-to-fine optimization scheme.

\paragraph{Loss weight scheduling.}
While we follow the individual loss objective terms and scheduling of NDS \citep{Worchel-CVPR-2022nds} for our mesh optimization loss in Sec. \textcolor{red}{3.2}, we added our side loss term $L_{\mathrm{sides}}$ to the objective with weight term $\lambda_{\mathrm{sides}}=0.1$, which we decrease by 10\% after each remeshing. We also set the loss weights for $L_{\mathrm{normal}}$ equivalent to $L_{\mathrm{shading}}$ in the original paper for NDS. During optimization, we progressively increase the geometric regularization term $L_{\mathrm{laplacian}}, L^{reg}_{\mathrm{normal}}$ to encourage the generation of smooth surfaces for the final mesh. For the second mesh refinement stage, which optimizes the earlier mesh based on the refined normal maps from multiple views (total of 36 views), we set $\lambda_{\mathrm{sides}} = 0$ since the side views can now be well constrained without the sidewise loss. 

\paragraph{Refine by resampling.}
To refine the mesh from dual normal map-based optimization, we render both full body and face normal maps and refine them with resampling technique (Sec. \textcolor{red}{3.3}). Here, we render $36$-view normal maps with $10^{\circ}$ yaw interval, and set $(t_0, K)$ to $(0.02, 2)$, respectively, both for body and face normal map refinement.

\section{Qualitative Results}
\label{sec:supp_qualitative}

\begin{figure*}[!htbp] \centering 
    \includegraphics[width=\linewidth,]{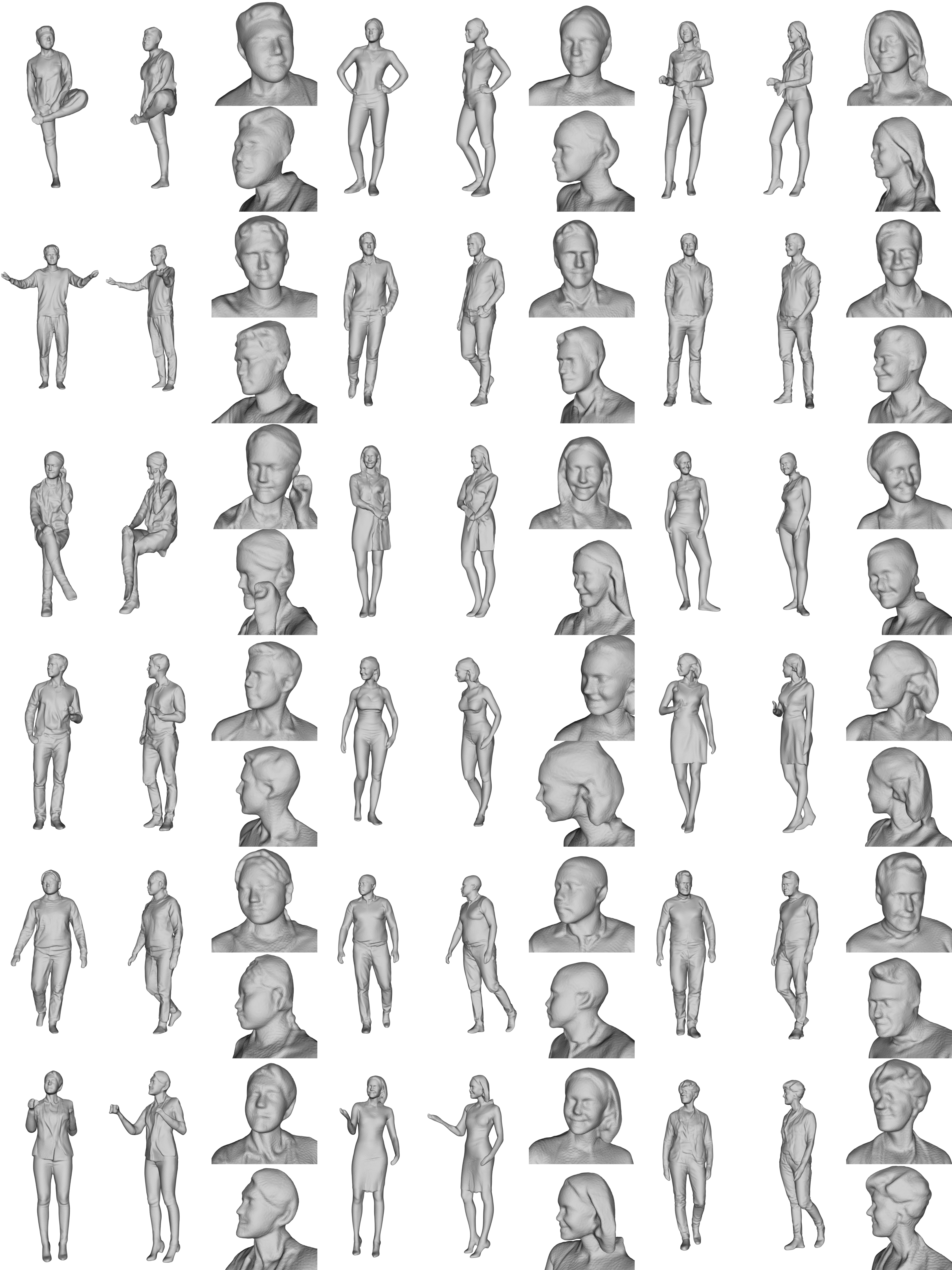}
    \caption{\textbf{More random generation results.}}
    \label{fig:more_random}
\end{figure*}

\paragraph{More generation results.}
\figref{fig:more_random} shows more random generation results from Chupa. We generate the human meshes based on SMPL-X parameters from the AGORA dataset~\cite{patel2021agora}, which includes SMPL, SMPL-X parameters fitted to $4,240$ 3D human scans. We can generate human scans with various identities and can be generalized to diverse poses.

\paragraph{Changing shape parameter $\pmb{\beta}$.}
To control the shape of the generated mesh, we can control the shape parameter $\pmb{\beta}$ of input SMPL-X mesh~\cite{Loper-TOG-2015smpl, Pavlakos-CVPR-2019smplx}. \figref{fig:beta0}, \figref{fig:beta1} shows the generated meshes according to the variation of $\pmb{\beta}$ with fixed pose parameter $\pmb{\theta}$, where $\beta_{1}$, $\beta_{2}$ corresponds to the first and second component of the shape parameter respectively~\cite{Loper-TOG-2015smpl}. 
\begin{figure*}[!htb]
     \centering
     \begin{subfigure}[b]{0.16\textwidth}
         \centering
         \includegraphics[width=\textwidth]{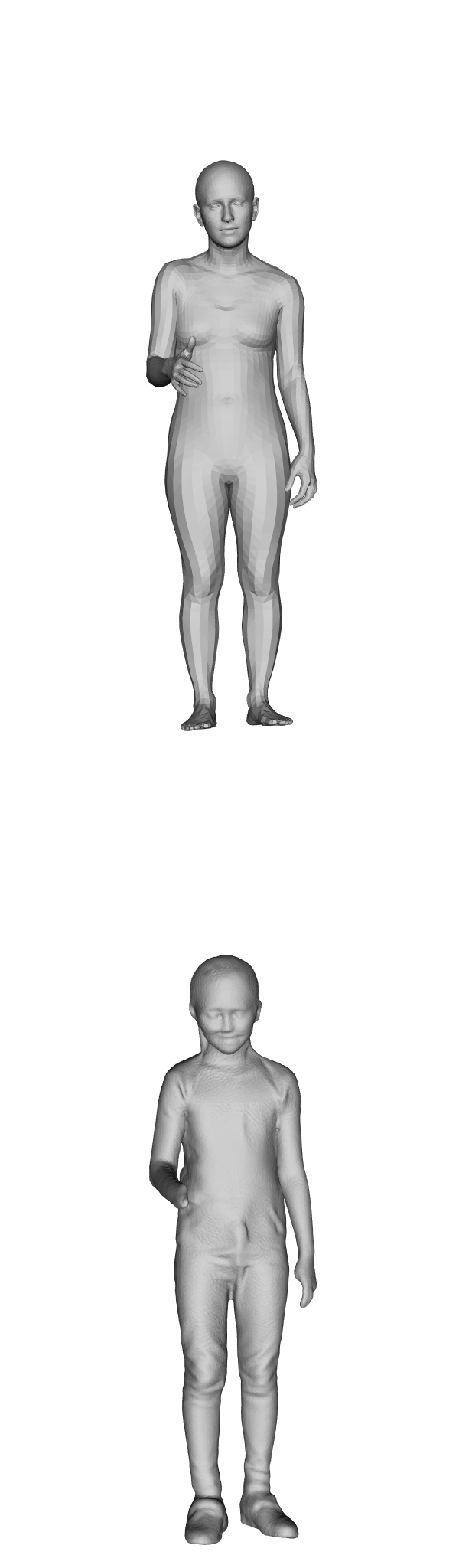}
         \caption{$\beta_{1}=-2$}
         \label{fig:beta0_-2}
     \end{subfigure}
     \begin{subfigure}[b]{0.16\textwidth}
         \centering
         \includegraphics[width=\textwidth]{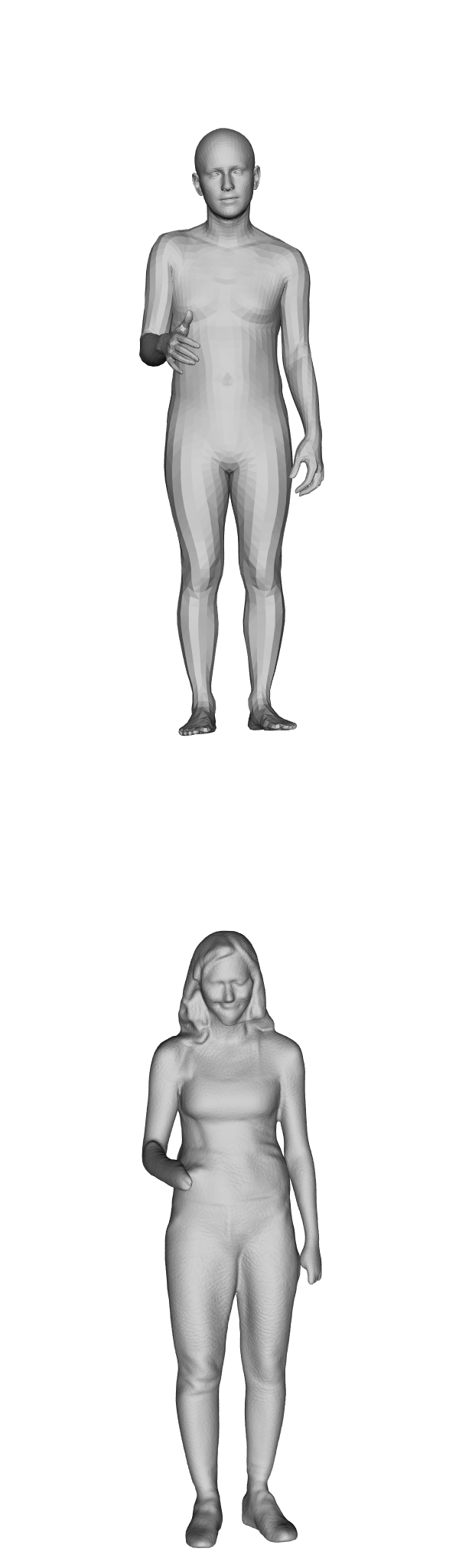}
         \caption{$\beta_{1}=-1$}
         \label{fig:beta0_-1}
     \end{subfigure}
     \begin{subfigure}[b]{0.16\textwidth}
         \centering
         \includegraphics[width=\textwidth]{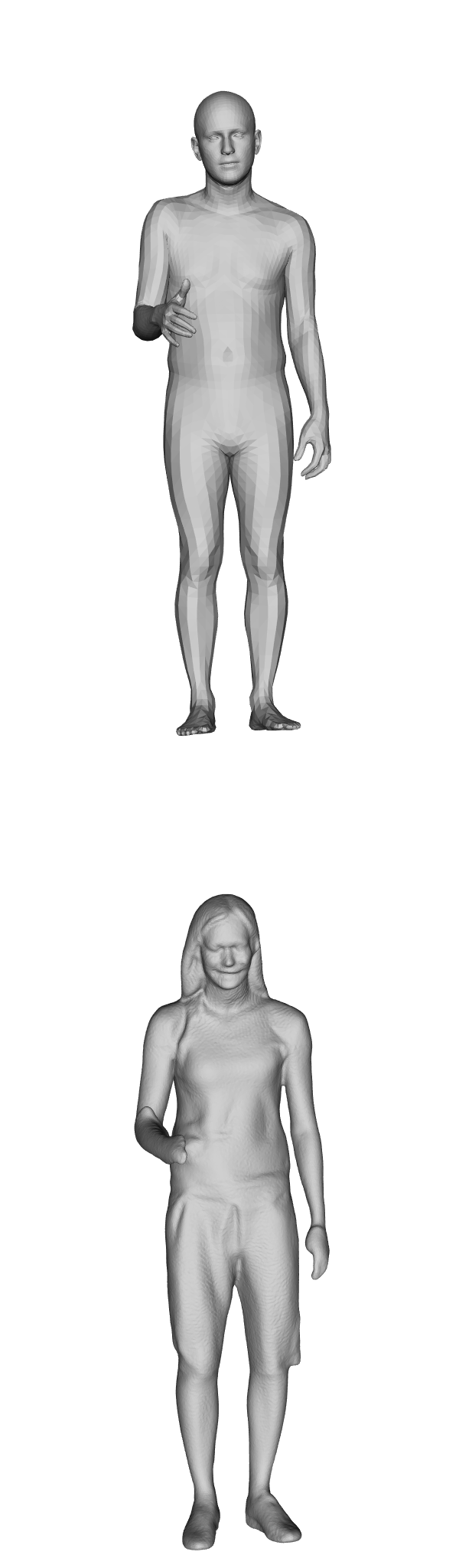}
         \caption{$\beta_{1}=0$}
         \label{fig:beta0_0}
     \end{subfigure}
     \begin{subfigure}[b]{0.16\textwidth}
         \centering
         \includegraphics[width=\textwidth]{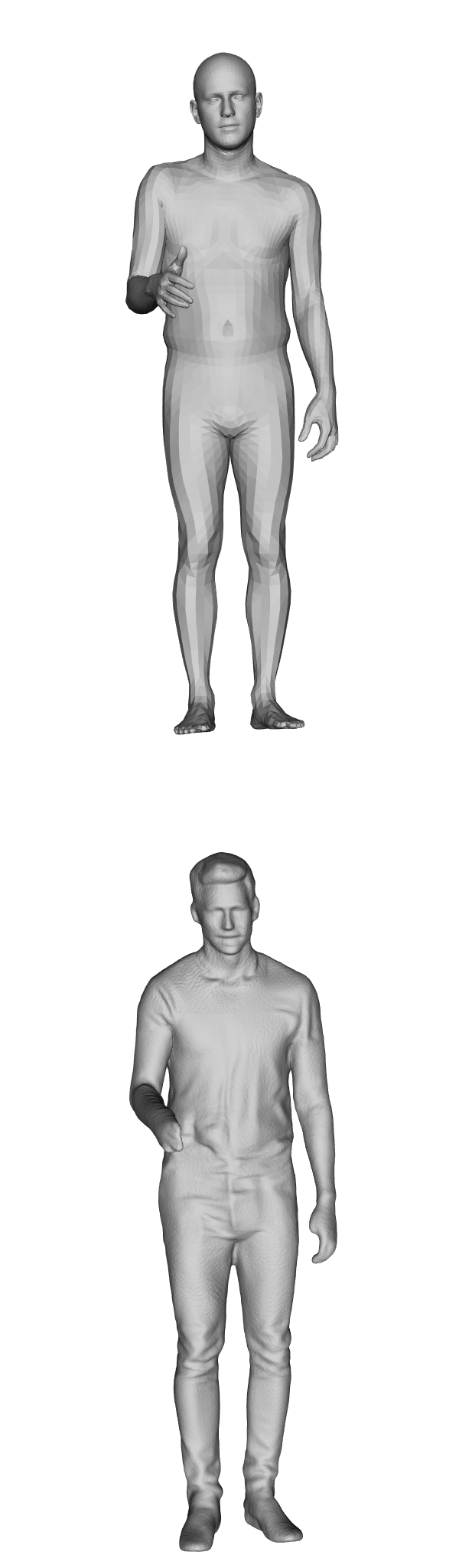}
         \caption{$\beta_{1}=1$}
         \label{fig:beta0_1}
     \end{subfigure}
     \begin{subfigure}[b]{0.16\textwidth}
         \centering
         \includegraphics[width=\textwidth]{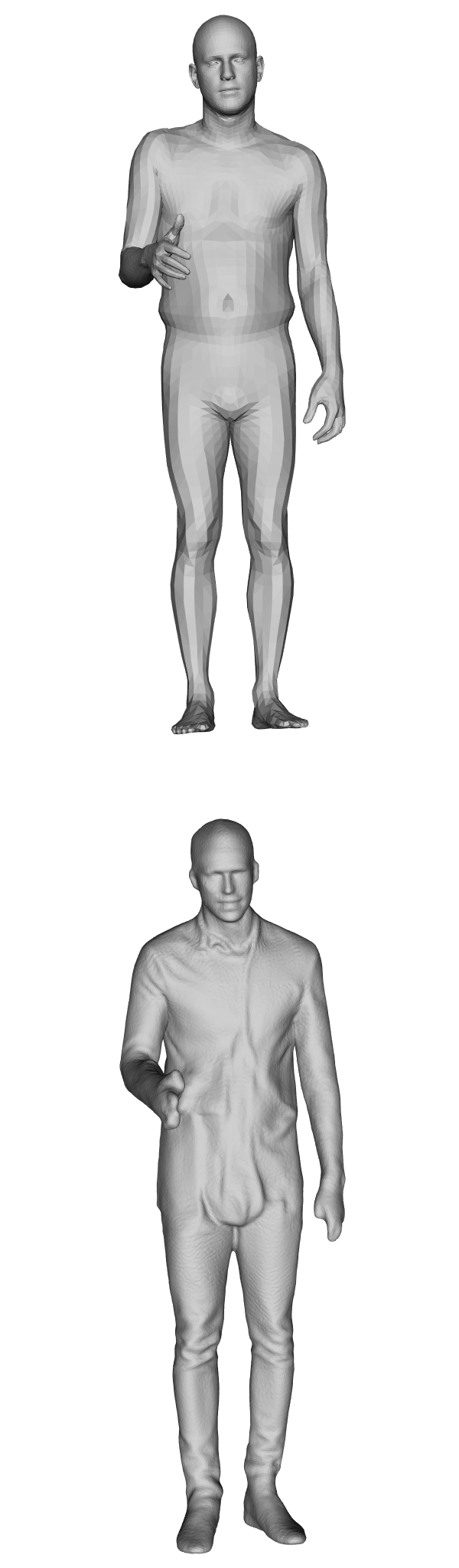}
         \caption{$\beta_{1}=2$}
         \label{fig:beta0_2}
     \end{subfigure}
     \hfill
    \caption{\textbf{Changing shape parameter $\pmb{\beta}_{1}$.}} 
\label{fig:beta0}
\end{figure*}

\begin{figure*}[!htb]
     \centering
     \begin{subfigure}[b]{0.16\textwidth}
         \centering
         \includegraphics[width=\textwidth]{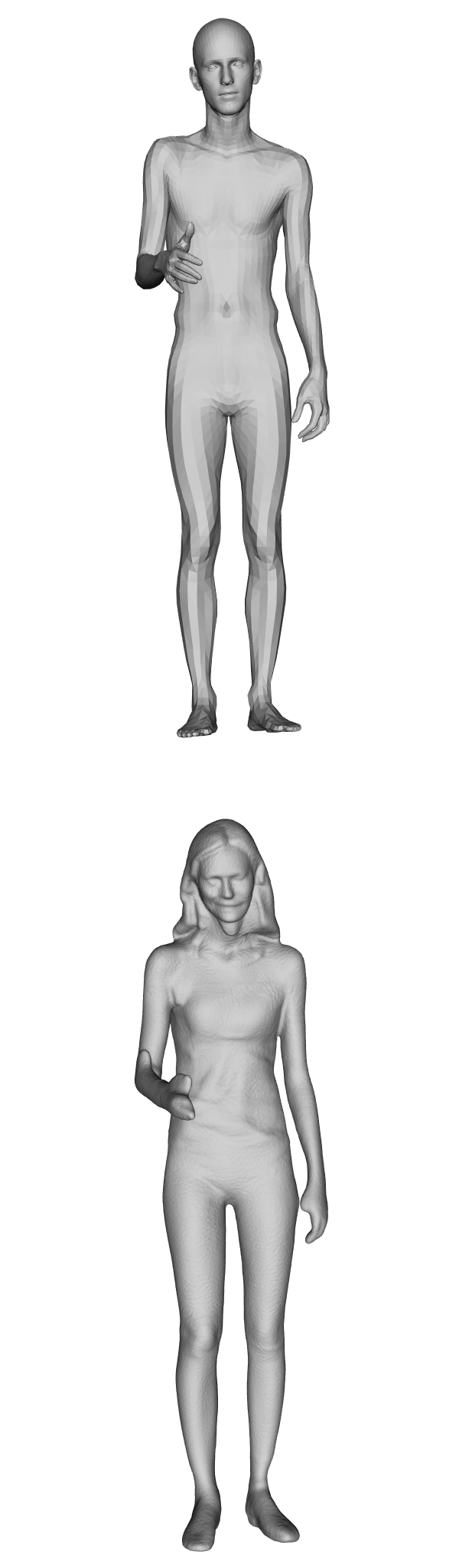}
         \caption{$\beta_{2}=-2$}
         \label{fig:beta1_-2}
     \end{subfigure}
     \begin{subfigure}[b]{0.16\textwidth}
         \centering
         \includegraphics[width=\textwidth]{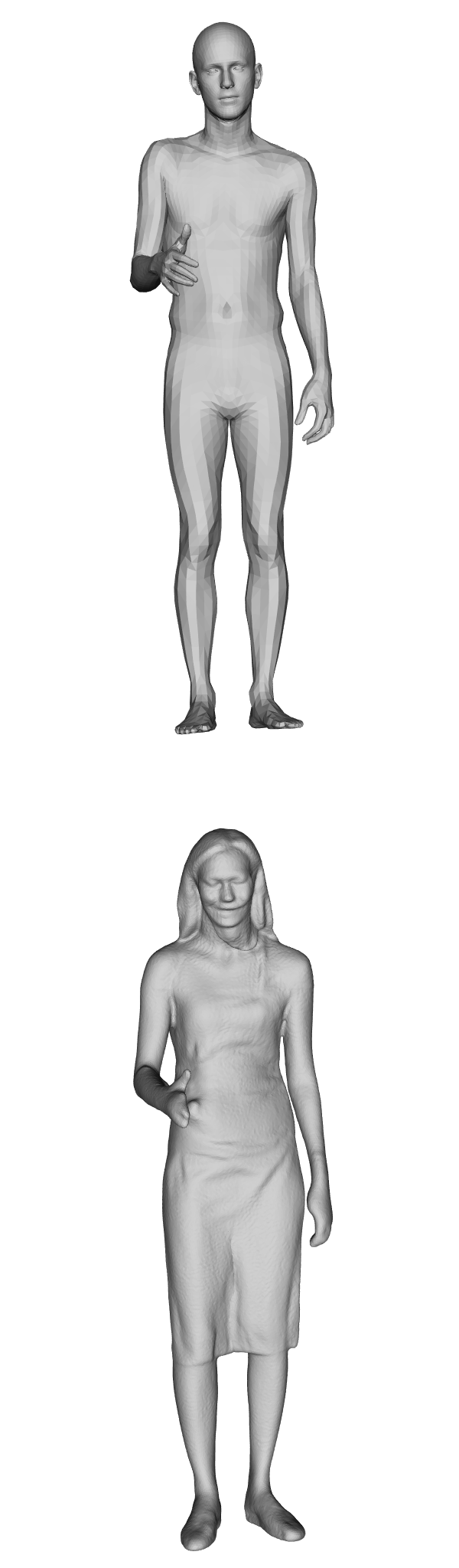}
         \caption{$\beta_{2}=-1$}
         \label{fig:beta1_-1}
     \end{subfigure}
     \begin{subfigure}[b]{0.16\textwidth}
         \centering
         \includegraphics[width=\textwidth]{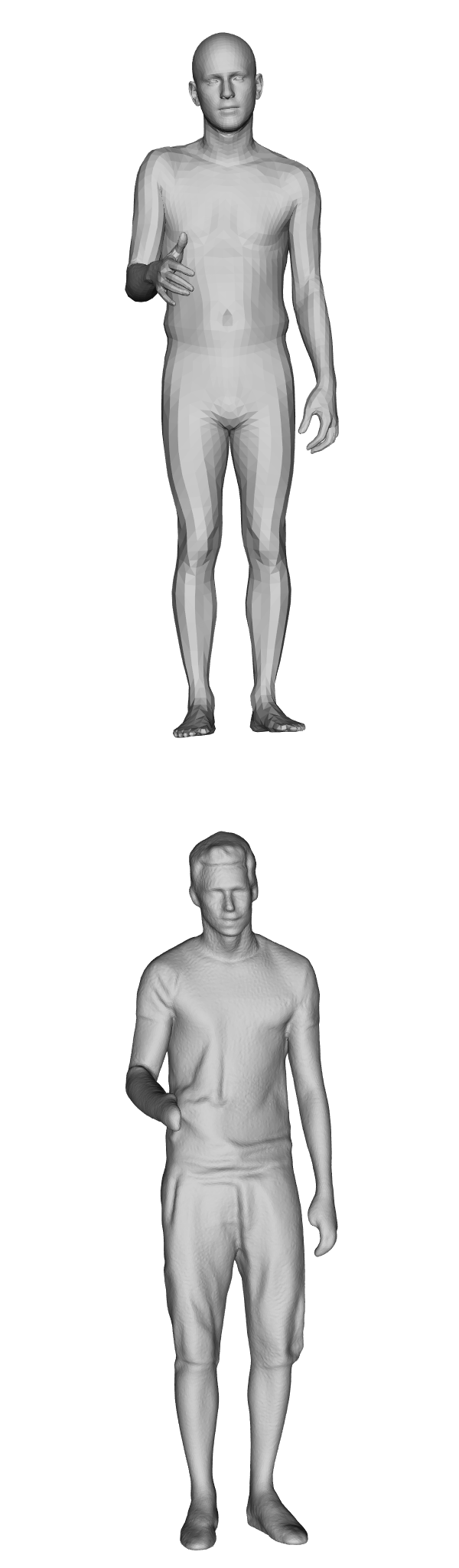}
         \caption{$\beta_{2}=0$}
         \label{fig:beta1_0}
     \end{subfigure}
     \begin{subfigure}[b]{0.16\textwidth}
         \centering
         \includegraphics[width=\textwidth]{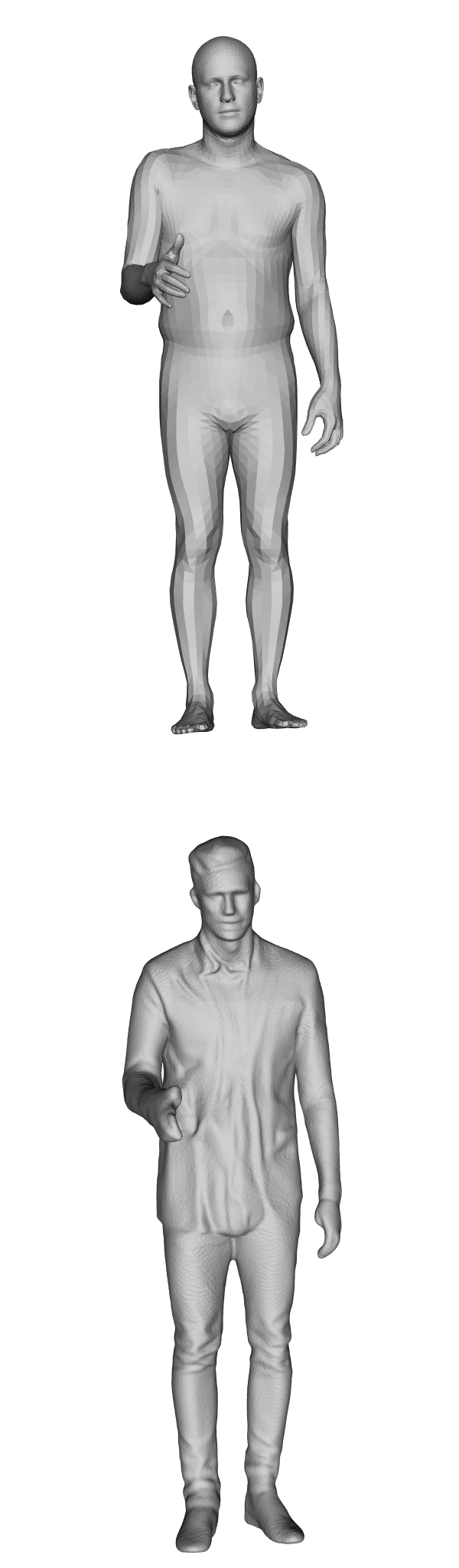}
         \caption{$\beta_{2}=1$}
         \label{fig:beta1_1}
     \end{subfigure}
     \begin{subfigure}[b]{0.16\textwidth}
         \centering
         \includegraphics[width=\textwidth]{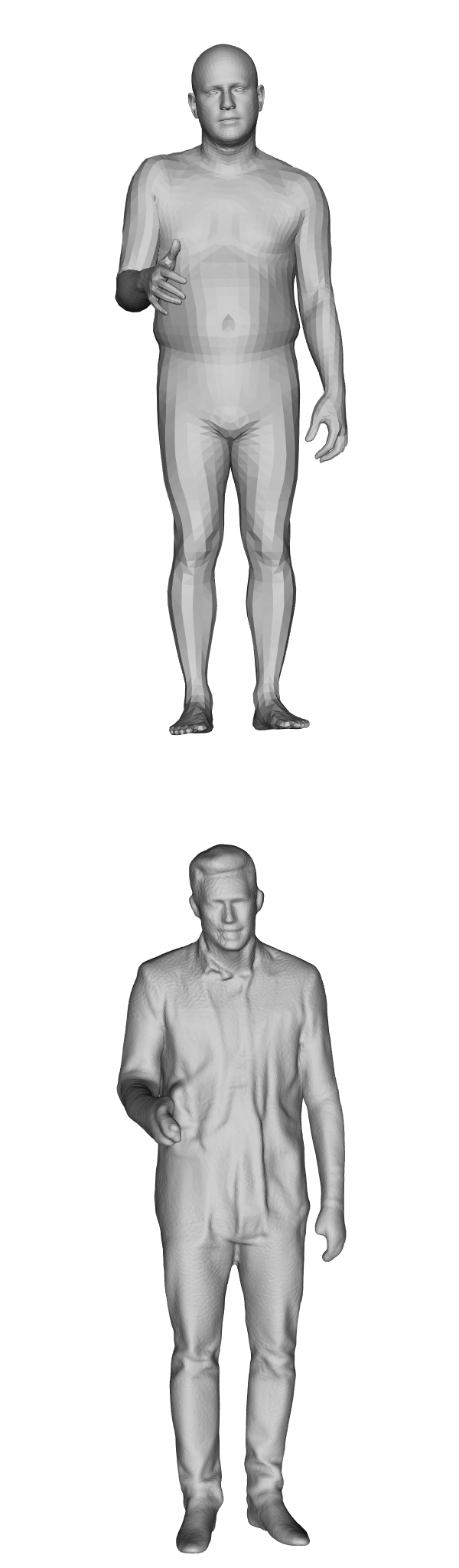}
         \caption{$\beta_{2}=2$}
         \label{fig:beta1_2}
     \end{subfigure}
     \hfill
    \caption{\textbf{Changing shape parameter $\pmb{\beta}_{2}$.}} 
\label{fig:beta1}
\end{figure*}

\begin{figure}[!htbp]
     \centering
     \begin{subfigure}[b]{0.9\columnwidth}
         \centering
         \includegraphics[width=\textwidth]{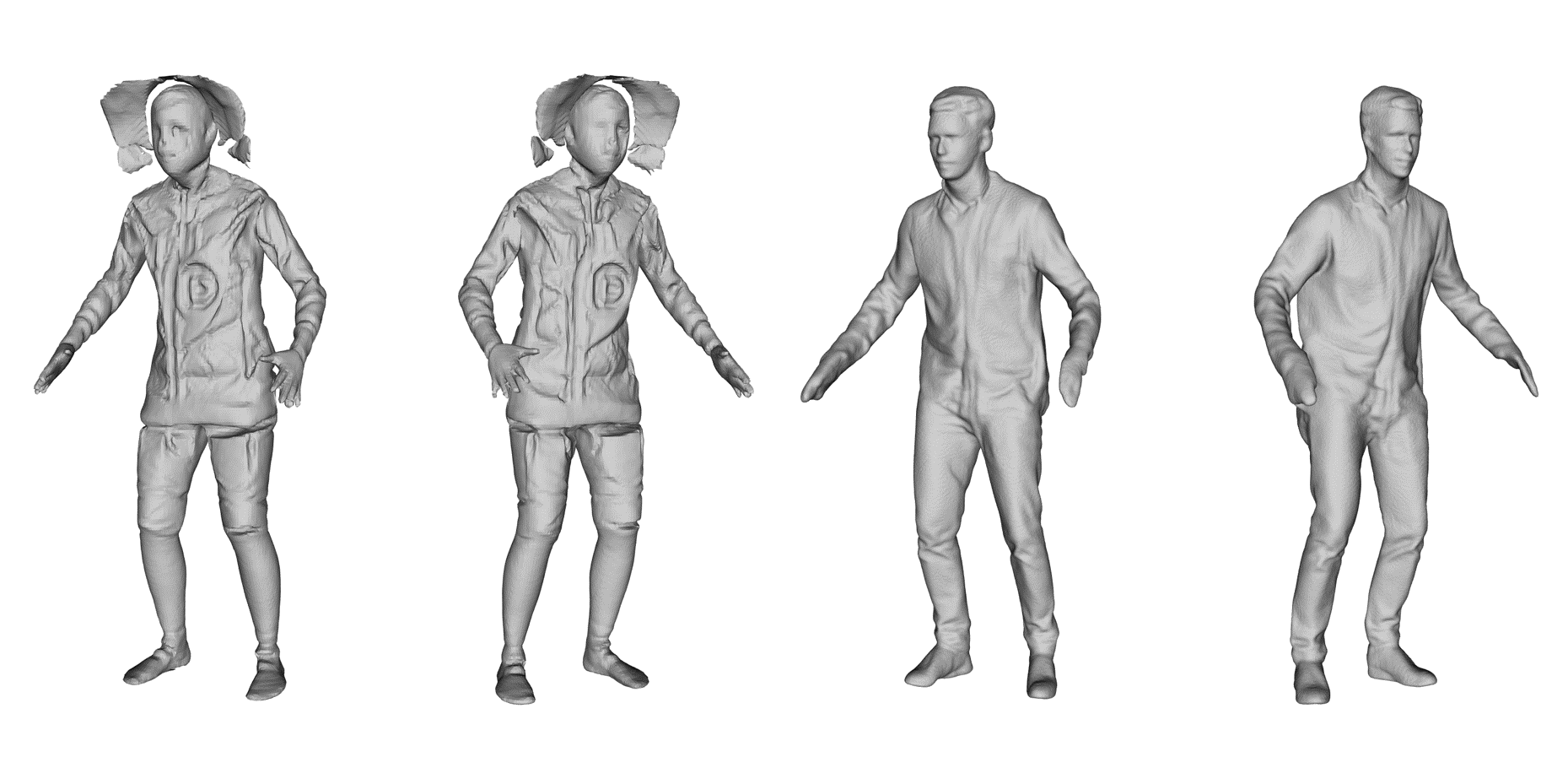}
         \caption{``a boy wearing a jacket''}
         \label{fig:a_boy_wearing_jacket}
     \end{subfigure}
     \begin{subfigure}[b]{0.9\columnwidth}
         \centering
         \includegraphics[width=\textwidth]{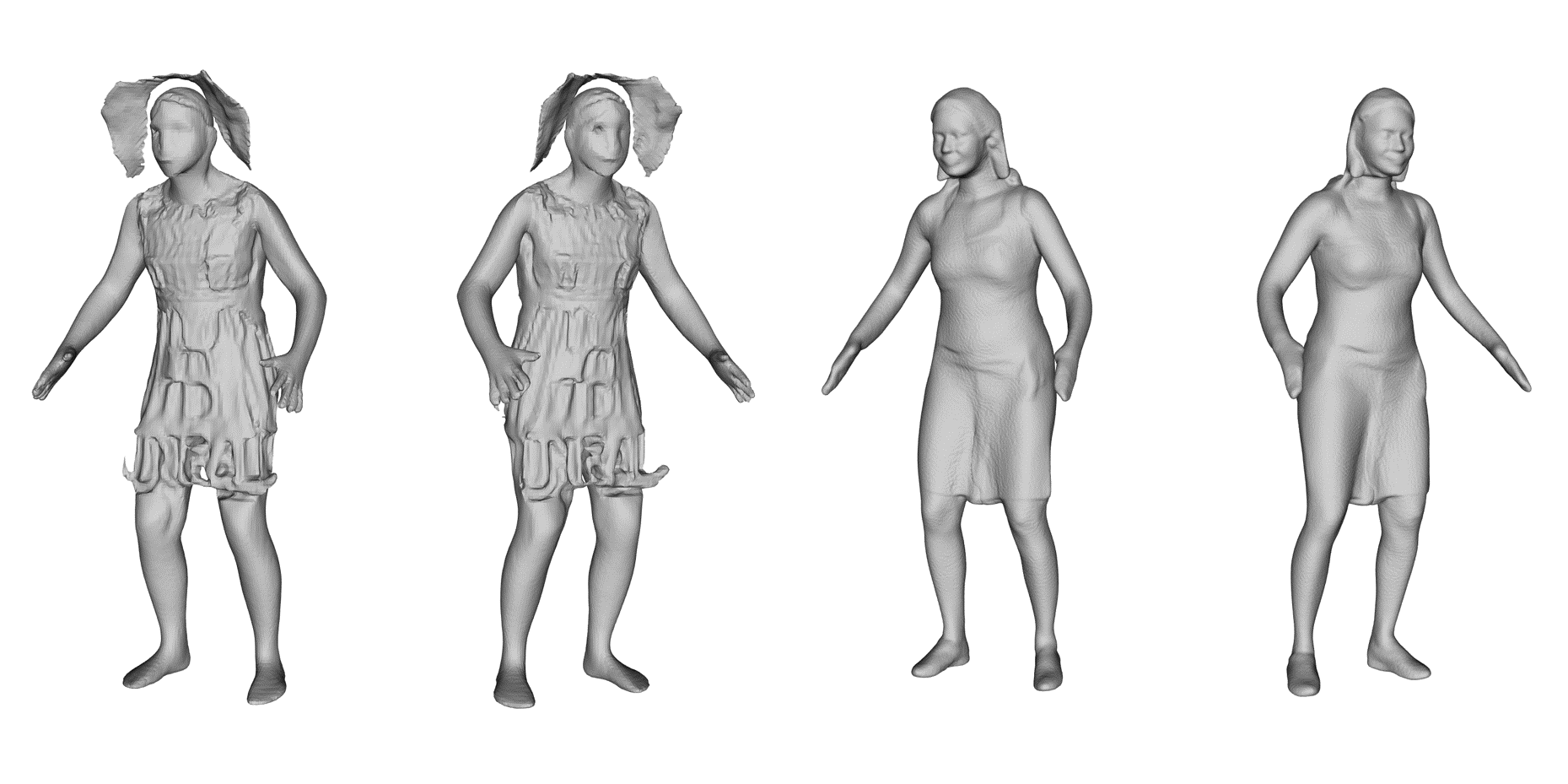}
         \caption{``a girl wearing a dress''}
         \label{fig:a_girl_wearing_dress}
     \end{subfigure}
     \hfill
    \caption{\textbf{Comparison with AvatarCLIP.} The left two columns are from AvatarCLIP, the right two columns are from Chupa (ours).} 
\label{fig:avatarclip}
\end{figure}

\begin{figure}[!htbp]
    \centering 
    \includegraphics[width=0.9\columnwidth]{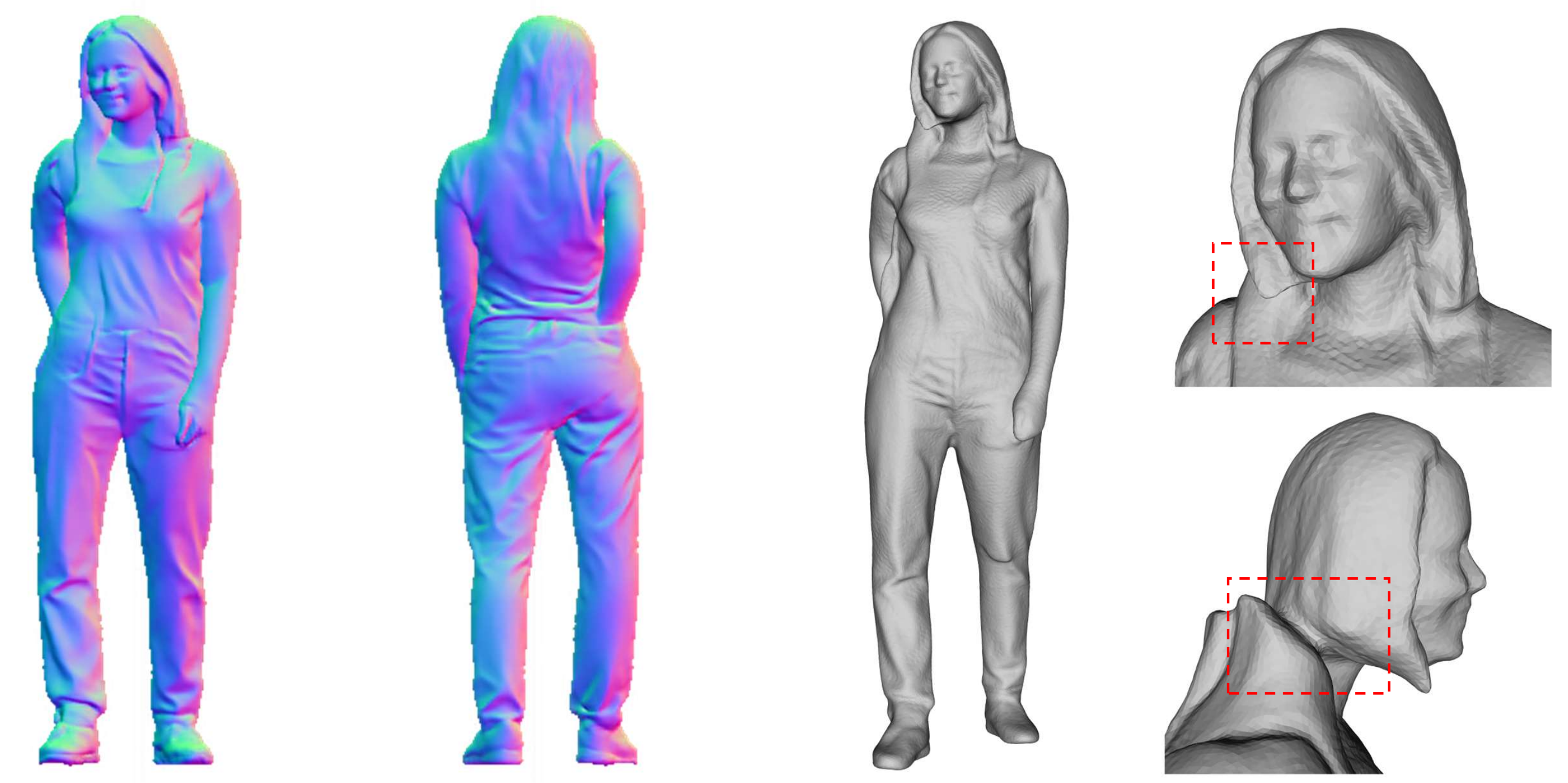}
    \caption{\textbf{Depth ambiguity problem.} Chupa may generate broken geometry, due to the depth ambiguity problem of our mesh reconstruction method(left: dual normal map, right: final mesh).}
    \label{fig:depth_ambiguity}
\end{figure}

\begin{figure}[!htbp]
    \centering 
    \includegraphics[width=0.9\columnwidth]{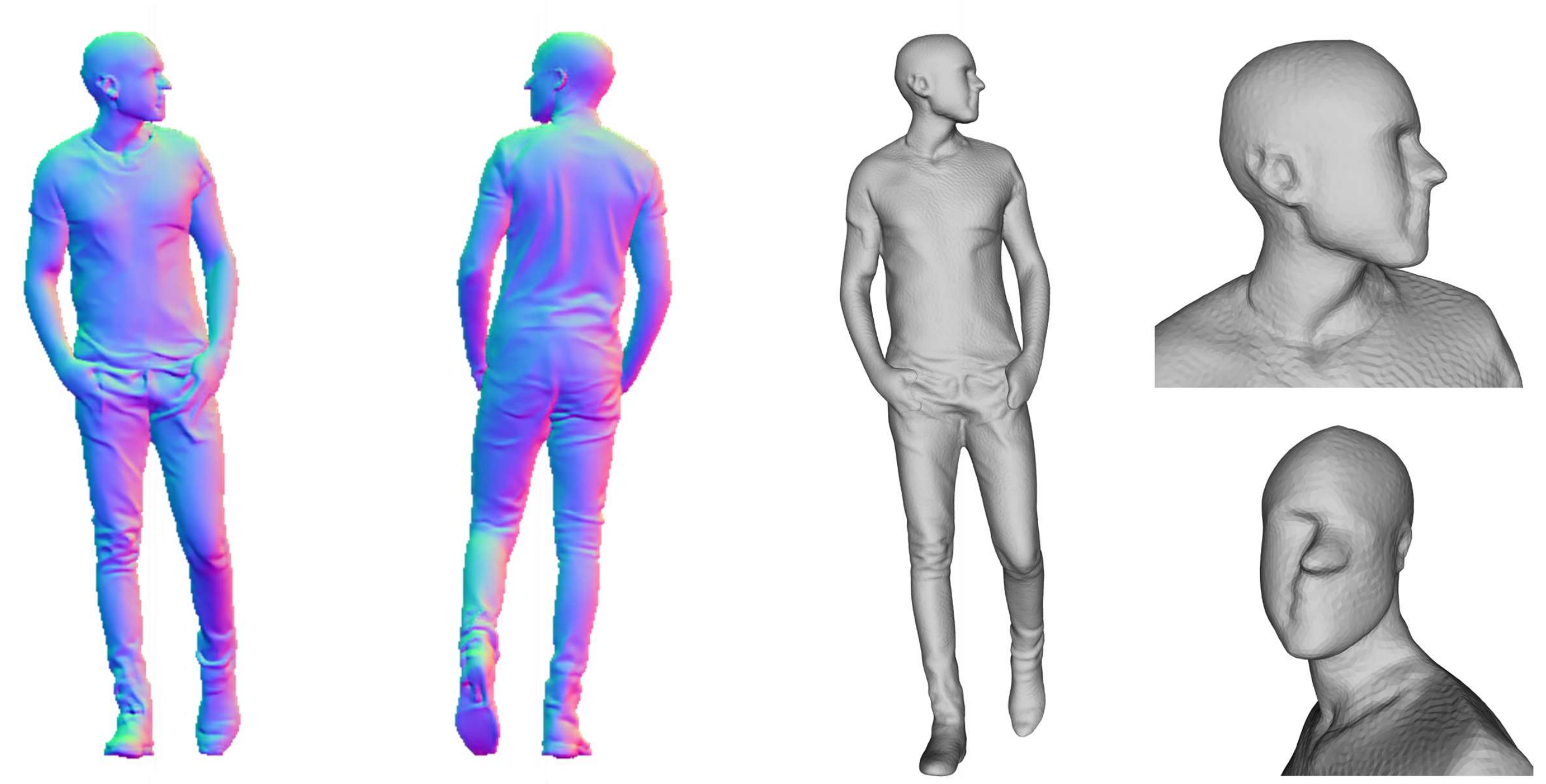}
    \caption{\textbf{Face direction matters.} Chupa may generate unnatural face geometry, when the face direction is not aligned with the input view (left: dual normal map, right: final mesh).}
    \label{fig:face_direction}
\end{figure}

\begin{figure}[!htbp]
    \centering 
    \includegraphics[width=0.9\columnwidth]{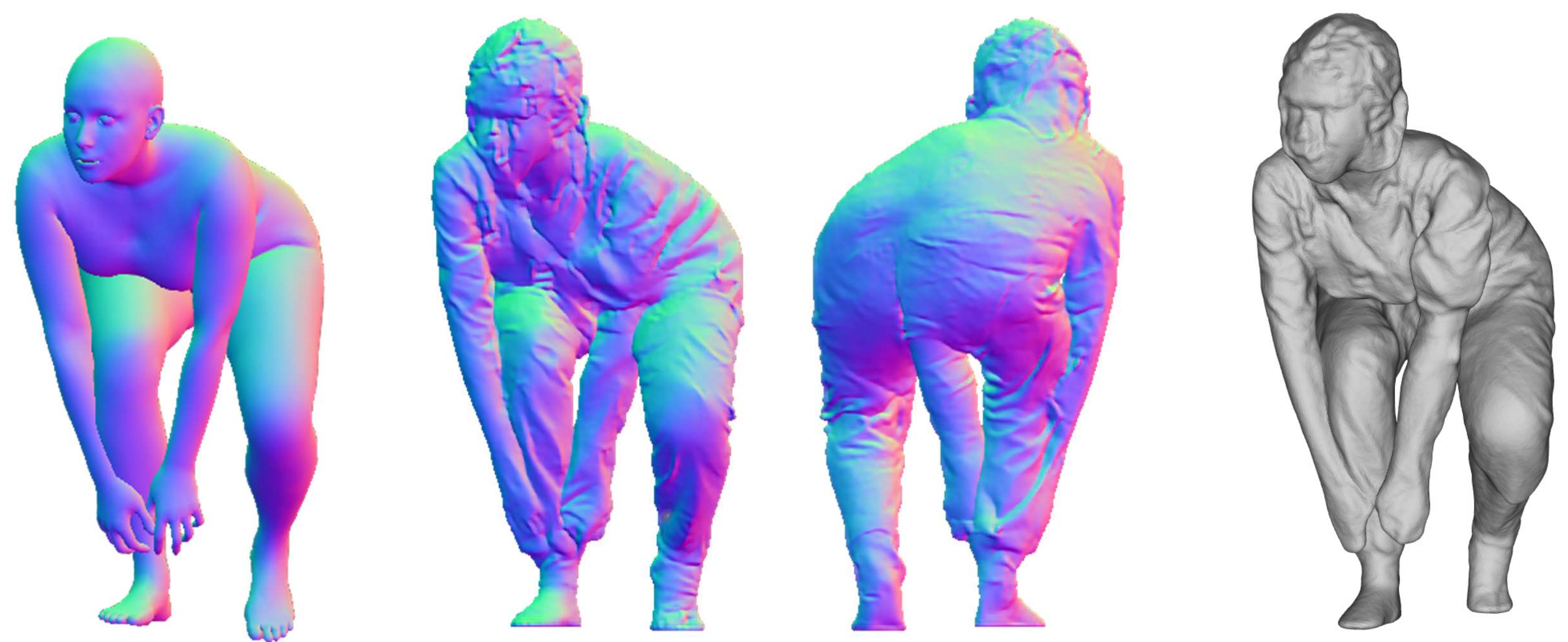}
    \caption{\textbf{Out-of-distribution pose.} Chupa may generate implausible geometry for some out-of-distribution pose (left: SMPL-X, middle: dual normal map, right: final mesh).}
    \label{fig:ood_pose}
\end{figure}

\paragraph{Comparison with AvatarCLIP.}
We compare our text-guided generation results with AvatarCLIP~\cite{hong-ACMTOG-2022avatarclip}, a text-guided 3D avatar generation pipeline that also initializes its 3D implicit surface model~\cite{wang2021neus} with a SMPL model. Once initialized, AvatarCLIP optimizes the 3D model based on a CLIP loss~\cite{radford-ICML2021-clip} on the rendered results, to match the 3D model according to the text description. \figref{fig:avatarclip} shows that Chupa can generate more realistic 3D human mesh while minimizing unnatural artifacts. Note that while AvatarCLIP takes more than 3 hours to generate a mesh, Chupa takes $3$ minutes with a single RTX3090.

\section{Failure Cases}
\label{sec:failure_cases}

\paragraph{Depth ambiguity problem.}
Our dual normal map-based mesh reconstruction method (Sec. \textcolor{red}{3.2}) has inherent depth ambiguity issues, as it only uses front and back-view normal maps for the initial optimization. When the given normal maps largely deviates from the initial SMPL model, \eg, long hair, the vertices for both head and shoulder deforms to match the provided hairstyle, creating artifacts during deformation. \figref{fig:depth_ambiguity} shows that while the hairstyle seems to be well-reconstructed in the front view, there exists unnatural seams and broken geometry at close view.

\paragraph{Face direction matters.} 

When the input pose contains misaligned body and face direction, the final output might display unnatural face geometry. For example, when the face is turned to the side direction (\figref{fig:face_direction}), the diffusion models might fail to generate realistic faces for reconstruction. To make matters worse, the small distortion due to depth ambiguity during reconstruction (Sec. \textcolor{red}{3.2}) can have huge impact on the perceptual quality of faces. \figref{fig:face_direction} shows an example of such cases, where the resulting face mesh displays unnatural geometry.

\paragraph{Out-of-distribution pose.}
While our method can be generalized for diverse poses, there exists out-of-distribution poses that the diffusion generative model fails to create plausible normal maps from. \figref{fig:ood_pose} shows such examples of unrealistic normal maps, which leads to 3D meshes with bad geometry.

\section{User Study}
\label{sec:supp_user_study}
We conduct a perceptual study asking user preference between the meshes from our method and gDNA~\cite{Chen-CVPR-2022gdna}. We collect $100$ participants through CloudResearch Connect~\cite{connect} and get $78$ valid answers out of them. Each participants are given $40$ problems which consist of $20$ problems for body and $20$ problems for face. \figref{fig:user_study} shows the example problems.

\begin{figure}[htbp]
     \centering
     \begin{subfigure}[b]{\columnwidth}
         \centering
         \includegraphics[width=\textwidth]{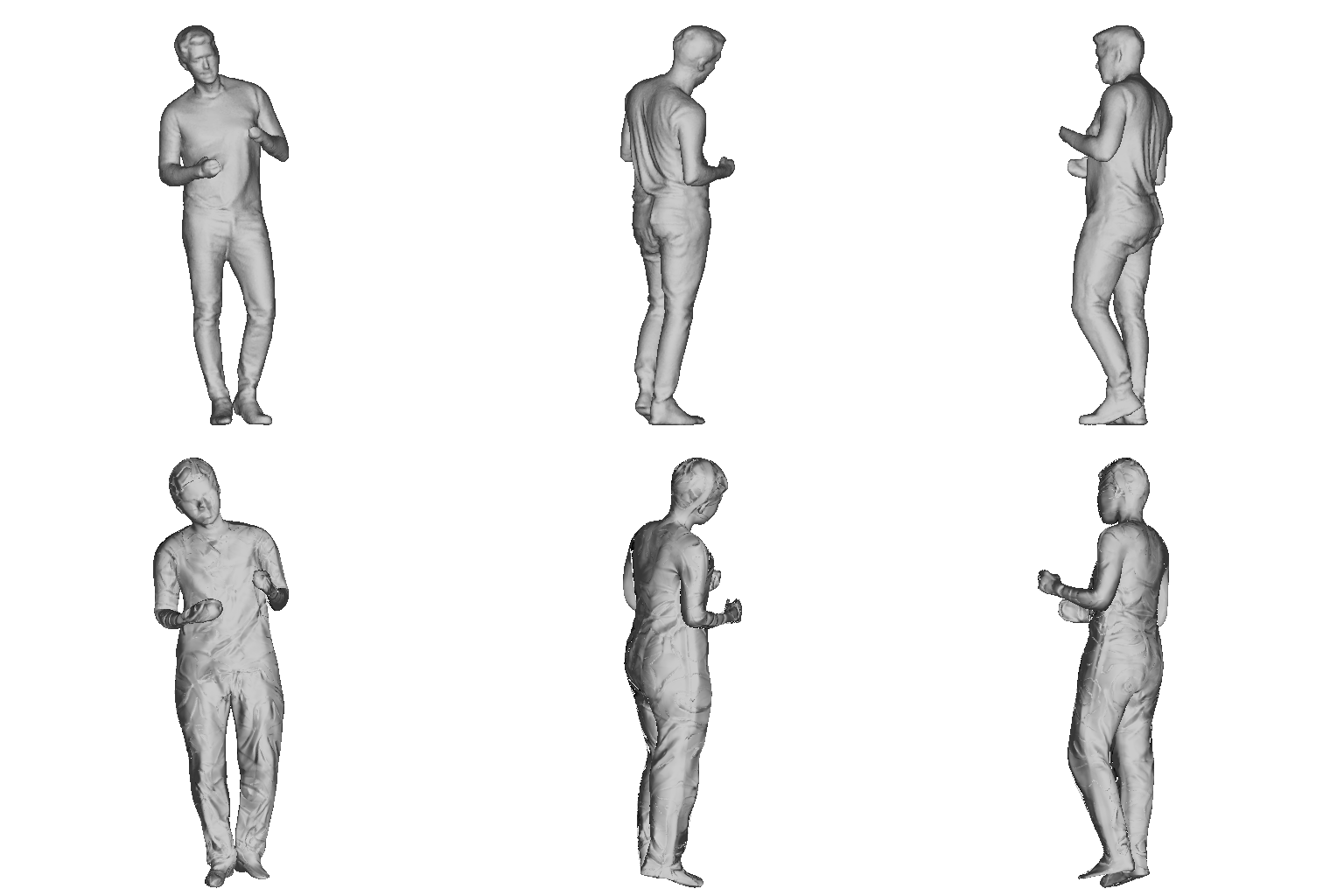}
         \caption{Body}
         \label{fig:users_study_body}
     \end{subfigure}
     \begin{subfigure}[b]{\columnwidth}
         \centering
         \includegraphics[width=\textwidth]{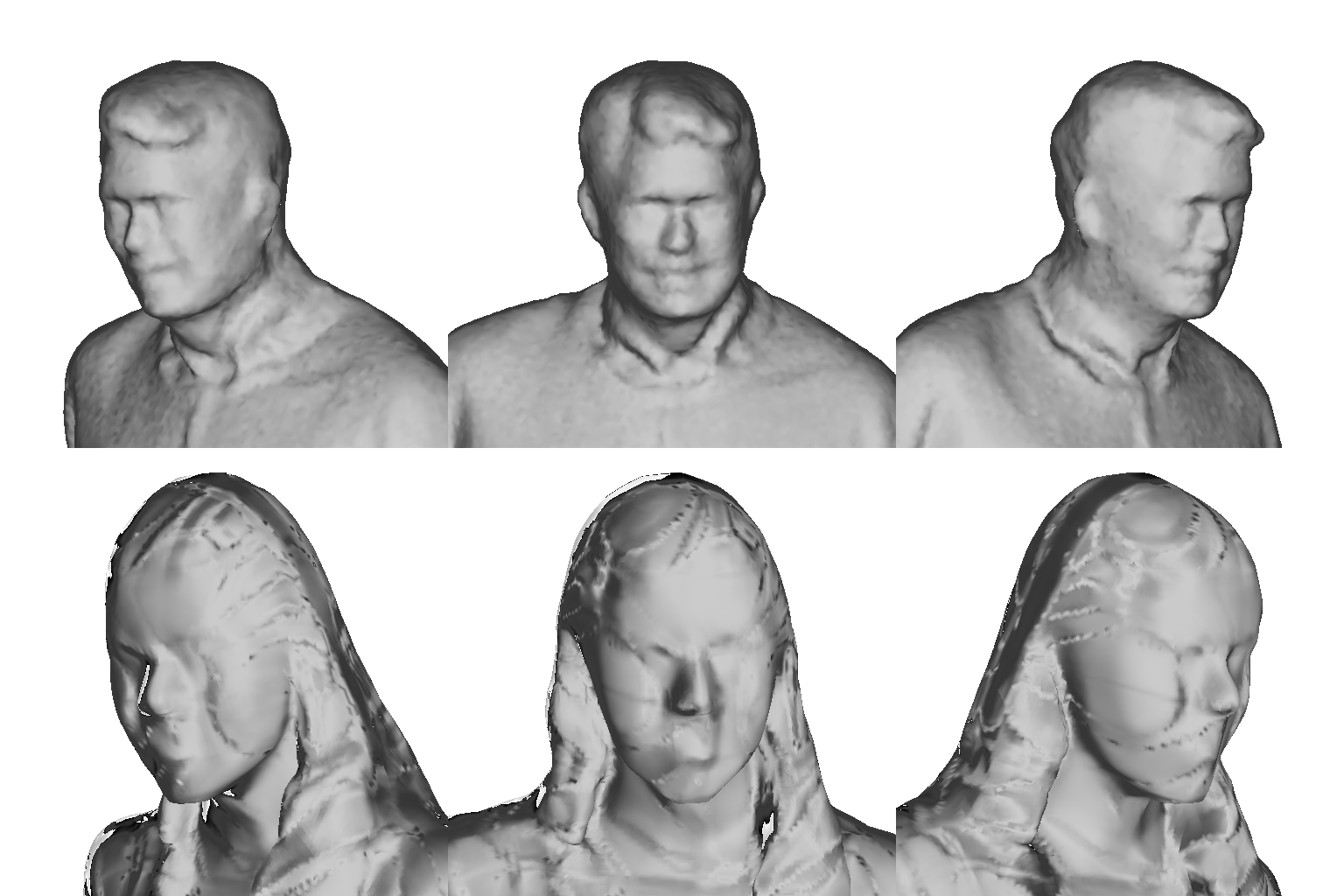}
         \caption{Face}
         \label{fig:users_study_face}
     \end{subfigure}
     \hfill
    \caption{\textbf{User study problem example}. The $3$ views of mesh from our method and gDNA~\cite{Chen-CVPR-2022gdna} with the same SMPL parameter are rendered as shading images. Each user is asked to choose more realistic shapes between two rows, where each row corresponds to the images from each method. Two rows are randomly shuffled.}
\label{fig:user_study}
\end{figure}

\section{Ablation Study}
\label{sec:supp_ablation}

We present additional ablation study results on changing various hyperparameters such as resampling parameters, sampling angle, and the sampling scheme for dual generation. In \tabref{tab:supp_refine} and \tabref{tab:supp_angle}, we present the effect of choosing different refinement parameters $(t_0, K)$ and the sampling angle during the refinement stage for both shaded and normal maps of the resulting meshes. We also present the effect of using different diffusion samplers in \tabref{tab:supp_sampling}.

\begin{table}[!htbp]
\caption{
\textbf{Ablation study on resampling.} We see the effects of $(t_{0}, K)$ both for body and face, with the number of views fixed as $36$.
}
\centering
\resizebox{0.8\columnwidth}{!}{
\begin{tabular}{cccc}
\toprule
\multicolumn{2}{c}{$(t_{0}, K)$} & & \\
Body & Face & $\textrm{FID}_{\textrm{normal}}$ $\downarrow$ & $\textrm{FID}_{\textrm{shade}}$ $\downarrow$ \\
\midrule
$(0.02, 2)$ & - & $22.61$ & $37.13$   \\
$(0.02, 4)$ & - & $26.68$ & $46.19$  \\
$(0.02, 6)$ & - & $31.39$ & $51.98$  \\
$(0.04, 2)$ & - & $27.02$ & $46.34$  \\
$(0.06, 2)$ & - & $31.71$ & $52.65$  \\
$(0.02, 2)$ & $(0.02, 2)$ & $\mathbf{21.90}$ & $\mathbf{36.58}$  \\
$(0.02, 2)$ & $(0.02, 4)$ & $22.42$ & $37.57$  \\
$(0.02, 2)$ & $(0.02, 6)$ & $22.65$ & $38.11$ \\
$(0.02, 2)$ & $(0.04, 2)$ & $22.41$ & $37.64$  \\
$(0.02, 2)$ & $(0.06, 2)$ & $22.65$ & $37.94$  \\
\bottomrule
\end{tabular}
}
\vspace{-0.05in}
\label{tab:supp_refine}
\end{table}
\begin{table}[!htbp]
\caption{
\textbf{Ablation on the number of views for refinement}. We see the effects of the number of views for refinement with $t_{0}=0.02, K=2$ as fixed.
} \centering
\begin{tabular}{llcc}
\toprule
$N_\mathrm{views}$ & $\theta_{\mathrm{step}}$ & FID$_{\mathrm{normal}}\downarrow$ & FID$_{\mathrm{shade}} \downarrow$ \\ \hline
\midrule
$4$ & $90^{\circ}$ & $30.88$ & $41.85$ \\
$6$ & $60^{\circ}$ & $29.01$ & $41.30$ \\
$12$ & $30^{\circ}$ & $25.21$ & $39.53$ \\
$36$ & $10^{\circ}$ & $\mathbf{21.90}$ & $\mathbf{36.58}$ \\
\bottomrule
\end{tabular}
\vspace{-0.05in}
\label{tab:supp_angle}
\end{table}
\begin{table}[!htbp]
\caption{
\textbf{Ablation on sampling scheme}. We ablate on the sampling scheme of our diffusion model for dual normal map generation. Here, we compute FID scores based on the results of dual normal map-based optimization without refinement.
} \centering
\begin{tabular}{lcc}
\toprule
Method  & FID$_{\mathrm{normal}}\downarrow$ & FID$_{\mathrm{shade}}\downarrow$\\ \hline
\midrule
Euler~\cite{Karras-NeurIPS-2022EDM}  & $28.84$ & $37.36$\\
DDIM~\cite{Song-ICLR-2021DDIM} & $26.76$ & $34.79$ \\
DDPM~\cite{Ho-NIPS-2020ddpm} & $26.31$ & $37.13$ \\
\bottomrule
\end{tabular}
\vspace{-0.05in}
\label{tab:supp_sampling}
\end{table}

\paragraph{Refine by resampling.}
\tabref{tab:supp_refine} shows the effects of varying $(t_0, K)$ for resampling. The first $6$ rows show the results of varying $(t_0, K)$ for body normal map refinement without face refinement. And the next $6$ rows show the results of varying $(t_0, K)$ for face normal map refinement with fixed $(t_0, K)$ for body normal map refinement. For both body and face, the smaller forward time steps and fewer iterations show better performance since large forward steps or many iterations may lead to the normal map inconsistent with the original normal maps.

\paragraph{The number of views for mesh refinement.}
\tabref{tab:supp_angle} shows the performance with the varying number of views used for the mesh refinement stage (Sec. \textcolor{red}{3.3}), where $N_{\mathrm{views}}, \theta_{\mathrm{step}}$ correspond to the number of views and the yaw interval between views respectively. Here, the hyperparameters $(t_0, K)$ for resampling are fixed as $(0.02, 2)$. It shows that increasing the number of views leads to better performance.

\paragraph{Sampling scheme of the diffusion model.}
As mentioned in Sec. \textcolor{red}{4.1}, we generate dual normal maps with the same denoising steps used during training, which is the sampling scheme of DDPM~\cite{Ho-NIPS-2020ddpm}. Here, we ablate on the different sampling schemes for diffusion probabilistic models, with two additional samplers~\cite{Song-ICLR-2021DDIM, Karras-NeurIPS-2022EDM} set to $t = 50$. \tabref{tab:supp_sampling} shows that the sampling scheme doesn't affect the performance significantly. Note that we compute the score without the mesh refinement stage (Sec. \textcolor{red}{3.3}) to analyze the effects of the sampler since the refinement stage only involves a small number of denoising steps. 

 \fi

\end{document}